\pdfoutput=1

\documentclass[11pt]{article}

\usepackage{ACL2023}

\usepackage{times}
\usepackage{latexsym}

\usepackage{graphicx} 
\graphicspath{ {./Images/} }

\usepackage{xcolor}
\usepackage{comment}
\usepackage{hyperref}
\usepackage{amsmath}
\usepackage{amssymb}
\usepackage{makecell}
\usepackage{algorithm}
\usepackage{algpseudocode}

\usepackage[T1]{fontenc}

\usepackage[utf8]{inputenc}

\usepackage{microtype}

\usepackage{inconsolata}

\title{Are Synonym Substitution Attacks Really \textit{Synonym} Substitution Attacks?}

\author{Cheng-Han Chiang \\
  National Taiwan University,\\ Taiwan\\
  \texttt{dcml0714@gmail.com} \\\And
   Hung-yi Lee \\
  National Taiwan University,\\ Taiwan \\
  \texttt{hungyilee@ntu.edu.tw} \\}

\begin{document}

\maketitle
\begin{abstract}
In this paper, we explore the following question: Are synonym substitution attacks really \textit{synonym} substitution attacks (SSAs)?
We approach this question by examining how SSAs replace words in the original sentence and show that there are still unresolved obstacles that make current SSAs generate invalid adversarial samples. 
We reveal that four widely used word substitution methods generate a large fraction of invalid substitution words that are ungrammatical or do not preserve the original sentence's semantics.
Next, we show that the semantic and grammatical constraints used in SSAs for detecting invalid word replacements are highly insufficient in detecting invalid adversarial samples.
\end{abstract}

\section{Introduction}
Deep learning-based natural language processing models have been extensively used in different tasks in many domains and have shown strong performance in different realms.
However, these models seem to be astonishingly vulnerable in that their predictions can be misled by some small perturbations in the original input~\citep{gao2018black,tan2020s}.
These \textit{imperceptible} perturbations, while not changing humans' predictions, can make a well-trained model behave worse than random.

One important type of adversarial attack in natural language processing (NLP) is the \textbf{synonym substitution attack} (SSA).
In SSAs, an adversarial sample is constructed by substituting some words in the original sentence with their synonyms~\citep{alzantot-etal-2018-generating,ren-etal-2019-generating, garg-ramakrishnan-2020-bae, jin2020bert,li2020bert,maheshwary2021strong}.
This ensures that the adversarial sample is semantically similar to the original sentence, thus fulfilling the imperceptibility requirement of a valid adversarial sample.
While substituting words with their semantic-related counterparts can retain the semantics of the original sentence, these attacks often utilize constraints to further guarantee that the generated adversarial samples are grammatically correct and semantically similar to the original sentence.
These SSAs have all been shown to successfully bring down well-trained text classifiers' performance.

However, some recent works observe, by human evaluations, that the quality of the generated adversarial samples of those SSAs is fairly low and is highly perceptible by human~\citep{morris-etal-2020-reevaluating, hauser2021bert}.
These adversarial samples often contain grammatical errors and do not preserve the semantics of the original samples, making them difficult to understand. 
These characteristics violate the fundamental criteria of a \textbf{\textit{valid adversarial sample}}: preserving semantics and being imperceptible to humans. 
This motivates us to investigate what causes those SSAs to generate invalid adversarial samples.
Only by answering this question can we move on to design more realistic SSAs in the future.

In this paper, we are determined to answer the following question: Are synonym substitution attacks in the literature really \textit{synonym} substitution attacks?
We explore the answer by scrutinizing the key components in several important SSAs and why they fail to generate valid adversarial samples.
Specifically, we conduct a detailed analysis of how the word substitution sets are obtained in SSAs, and we look into the semantic and grammatical constraints used to filter invalid adversarial samples.
We have the following astonishing observations:
\begin{itemize}
    \item When substituting words by WordNet synonym sets, current methods neglect the word sense differences within the substitution set. (Section~\ref{subsection:WordNet Synonym Set Substitution Ignores Word Senses})
    \item When using counter-fitted GloVe embedding space or BERT to generate the substitution set, the substitution set only contains a teeny-tiny fraction of synonyms. (Section~\ref{subsection:Counter-fitted Embedding $k$NN and BERT MLM/Reconstruction Contain Few Matched Sense Synonym})
    \item Using word embedding cosine similarity or sentence embedding cosine similarity to filter words in the substitution set does not necessarily exclude semantically invalid word substitutions. (Section~\ref{subsection:Valid Word Substitutions Do Not Necessarily Have Higher Word Embedding Cosine Similarity} and Section~\ref{subsection:Sentence Encoder Is Insensitive to Invalid Word Substitutions})
    \item The grammar checker used for filtering ungrammatical adversarial samples fails to detect most erroneous verb inflectional forms in a sentence. (Section~\ref{subsection:Language-tool Cannot Detect False Verb Inflectional Form})
\end{itemize}

\section{Backgrounds}
\label{section:Backgrounds}
In this section, we provide an overview of SSAs and introduce some related notations that will be used throughout the paper.

\subsection{Synonym Substitution Attacks (SSAs)}
\label{subsection:SSAs}
Given a victim text classifier trained on a dataset $D_{train}$ and a clean testing data $\mathbf{x}_{ori}$ sampled from the same distribution of $D_{train}$; $\mathbf{x}_{ori}=\{x_1,\cdots, x_{T}\}$ is a sequence with $T$ tokens. 
An SSA attacks the victim model by constructing an adversarial sample $\mathbf{x}_{adv}=\{x^{'}_1, \cdots, x^{'}_{T}\}$ by swapping the words in $\mathbf{x}_{ori}$ with their semantic-related counterparts.
For $\mathbf{x}_{adv}$ to be considered as a \textbf{valid} adversarial sample of  $\mathbf{x}_{ori}$, a few requirements must be met~\citep{morris-etal-2020-reevaluating}:
(0) $\mathbf{x}_{adv}$ should make the model yield a wrong prediction while the model can correctly classify $\mathbf{x}_{ori}$. 
(1) $\mathbf{x}_{adv}$ should be semantically similar with $\mathbf{x}_{ori}$. 
(2) $\mathbf{x}_{adv}$ should not induce new grammar errors compared with $\mathbf{x}_{ori}$.
(3) The word-level overlap between $\mathbf{x}_{adv}$ and $\mathbf{x}_{ori}$ should be high enough.
(4) The modification made in $\mathbf{x}_{adv}$ should be natural and non-suspicious.
\textbf{In our paper, we will refer to the adversarial samples that fail to meet the above criteria as invalid adversarial samples.}

SSAs rely on heuristic procedures to ensure that $\mathbf{x}_{adv}$ satisfies the preceding specifications.
Here, we describe a canonical pipeline of generating $\mathbf{x}_{adv}$ from $\mathbf{x}_{ori}$~\citep{morris2020textattack}.
Given a clean testing sample $\mathbf{x}_{ori}$ that the text classifier correctly predicts, an SSA will first generate a candidate word substitution set $\mathbb{S}_{x_i}$ for each word $x_i$.
The process of generating the candidate set $\mathbb{S}_{x_i}$ is called \textbf{transformation}.
Next, the SSA will determine which word in $\mathbf{x}_{ori}$ should be substituted first, and which word should be the next to swap, etc.
After the word substitution order is decided, the SSA will iteratively substitute each word $x_i$ in $\mathbf{x}_{ori}$ using the candidate words in $\mathbb{S}_{x_i}$ according to the pre-determined order.
In each substitution step, an $x_i$ is replaced by a word in $\mathbb{S}_{x_i}$, and a new $\mathbf{x}_{swap}$ is obtained.
When an $\mathbf{x}_{swap}$ is obtained, some \textbf{constraints} are used to verify the validity of $\mathbf{x}_{swap}$.
The iterative word substitution process will end if the model's prediction is successfully corrupted by a substituted sentence that sticks to the constraints, yielding the desired $\mathbf{x}_{adv}$ eventually.

Clearly, the transformations and the constraints are critical to the quality of the final $\mathbf{x}_{adv}$.
In the remaining part of the paper, we will look deeper into the transformations and constraints used in SSAs and their role in creating adversarial samples\footnote{In our paper, we do not discuss the relationship between the validity of an SSA and how an SSA determines which word in $\mathbf{x}_{ori}$ should be substituted. Most SSAs use word importance scores to determine what the most salient words are and substitute the most salient words. Since most SSAs use similar methods to determine what word should be replaced, our analyses are generalizable to those SSAs.}. 
Next, we briefly introduce the transformations and constraints that have been used in SSAs.

\subsection{Transformations}
\label{subsection:Transformation Methods}
Transformation is the process of generating the substitution set $\mathbb{S}_{x_i}$ for a word $x_i$ in $\mathbf{x}_{ori}$.
There are four representative transformations in the literature.

\paragraph{WordNet Synonym Transformation} constructs $\mathbb{S}_{x_i}$ by querying a word's synonym using WordNet~\citep{miller1995wordnet,princeton2010wordnet}, a lexical database containing the word sense definition, synonyms, and antonyms of the words in English.
This transformation is used in PWWS~\citep{ren-etal-2019-generating} and LexicalAT~\citep{xu-etal-2019-lexicalat}.

\paragraph{Word Embedding Space Nearest Neighbor Transformation} constructs $\mathbb{S}_{x_i}$ by looking up the word embedding of $x_i$ in a word embedding space, and finding its $k$ nearest neighbors ($k$NN) in the word embedding space.
Using $k$NN for word substitution is based on the assumption that semantically related words are closer in the word embedding space.
Counter-fitted GloVe embedding space~\citep{mrkvsic2016counter} is the embedding space obtained from post-processing the GloVe embedding space~\citep{pennington2014glove}.
Counter-fitting refers to the process of pulling away antonyms and narrowing the distance between synonyms.
This transformation is adopted in TextFooler~\citep{jin2020bert}, Genetic algorithm attack~\citep{alzantot-etal-2018-generating}, and TextFooler-Adj~\citep{morris-etal-2020-reevaluating}.

\paragraph{Masked Language Model (MLM) Mask-Infilling Transformation} constructs $\mathbb{S}_{x_i}$ by masking $x_i$ in $\mathbf{x}_{ori}$ and asking an MLM to predict the masked token; MLM's top-$k$ prediction of the masked token forms the word substitution set of $x_i$.
Widely adopted MLMs includes BERT~\citep{devlin2019bert} and RoBERTa~\citep{liu2019roberta}.
Using MLM mask-infilling to generate a candidate set relies on the belief that MLMs can generate fluent and semantic-consistent substitutions for $\mathbf{x}_{ori}$.
This method is used in BERT-ATTACK~\citep{li2020bert} and CLARE~\citep{li-etal-2021-contextualized}.

\paragraph{MLM Reconstruction Transformation} also uses MLMs.
When using MLM reconstruction transformation to generate the candidate set, one just feeds the MLM with the original sentence $\mathbf{x}_{ori}$ \textbf{without masking} any tokens in the sentence.
Here, the MLM is not performing mask-infilling but reconstructs the input tokens from the unmasked inputs.
For each word $x_i$, one can take its top-$k$ token reconstruction prediction as the candidates.
This transformation relies on the intuition that reconstruction can generate more semantically similar words than using mask-infilling.
This method is used in BAE~\citep{garg-ramakrishnan-2020-bae}.

\subsection{Constraints}
\label{subsection:Constraints}
When an $\mathbf{x}_{ori}$ is perturbed by swapping some words in it, we need to use some constraints to check whether the perturbed sentence, $\mathbf{x}_{swap}$, is semantically or grammatically valid or not.
We use $\mathbf{x}_{swap}$ instead of $\mathbf{x}_{adv}$ here as $\mathbf{x}_{swap}$ does not necessarily flip the model's prediction and thus not necessarily an adversarial sample.

\paragraph{Word Embedding Cosine Similarity}
requires a word $x_i$ and its perturbed counterpart $x^{'}_{i}$ to be close enough in the counter-fitted GloVe embedding space, in terms of cosine similarity.
A substitution is valid if its word embedding's cosine similarity with the original word's embedding is higher than a pre-defined threshold.
This is used in Genetic Algorithm Attack~\citep{alzantot-etal-2018-generating} and TextFooler~\citep{jin2020bert}.

\paragraph{Sentence Embedding Cosine Similarity}
demands that the sentence embedding cosine similarity between $\mathbf{x}_{swap}$ and $\mathbf{x}_{ori}$ are higher than a pre-defined threshold.
Most previous works~\citep{jin2020bert,li2020bert,garg-ramakrishnan-2020-bae,morris-etal-2020-reevaluating} use Universal Sentence Encoder (USE)~\citep{cer2018universal} as the sentence encoder; A2T~\citep{yoo2021towards} use a DistilBERT~\citep{sanh2019distilbert} fine-tuned on STS-B~\citep{cer2017semeval} as the sentence encoder.

In some previous work~\citep{li2020bert}, the sentence embedding is computed using the whole sentence $\mathbf{x}_{ori}$ and $\mathbf{x}_{swap}$.
But most previous works~\citep{jin2020bert,garg-ramakrishnan-2020-bae} only extract a context around the currently swapped word in $\mathbf{x}_{ori}$ and $\mathbf{x}_{swap}$ to compute the sentence embedding.
For example, if $x_i$ is substituted in the current substitution step, one will compute the sentence embedding between $\mathbf{x}_{ori}[i-w:i+w+1]$ and $\mathbf{x}_{adv}[i-w:i+w+1]$, where $w$ determines the window size.
$w$ is set to 7 in~\citet{jin2020bert} and~\citet{garg-ramakrishnan-2020-bae}.


\paragraph{\href{https://languagetool.org}{LanguageTool}}~\citep{languagetool} is an open-source grammar tool that can detect spelling errors and grammar mistakes in an input sentence.
It is used in TextFooler-Adj~\citep{morris-etal-2020-reevaluating} to evaluate the grammaticality of the adversarial samples.

\section{Problems with the Transformations in SSAs}
\label{subsection:Problems with Current Transformation Methods}
In this section, we show that the transformations introduced in Section~\ref{subsection:Transformation Methods} are largely to blame for the invalid adversarial samples in SSAs.
This is because the substitution set $\mathbb{S}_{x_i}$ for $x_i$ is mostly invalid, either semantically or grammatically.

\subsection{WordNet Synonym Substitution Set Ignores Word Senses}
\label{subsection:WordNet Synonym Set Substitution Ignores Word Senses}
In WordNet, each word is associated with one or more word senses, and each word sense has its corresponding synonym sets.
Thus, the substitution set $\mathbb{S}_{x_i}$ proposed by WordNet is the union of the synonym sets of different senses of $x_i$. 
When swapping $x_i$ with its synonym using WordNet, it is more sensible to first identify the word sense of $x_i$ in $\mathbf{x}_{ori}$, and use the synonym set of the very sense as the substitution set.
However, current attacks using WordNet synonym substitution neglect the sense differences within the substitution set~\citep{ren-etal-2019-generating}, which may result in adversarial samples that semantically deviate from the original input.

As a working example, consider a movie review that reads "I highly \textcolor{red}{recommend} it".
The word "recommend" here corresponds to the word sense of "\textit{express a good opinion of}" according to WordNet and has the synonym set \{recommend, commend\}.
Aside from the above word sense, "recommend" also have another word sense: "push for something", as in "The travel agent recommends not to travel amid the pandemic".
This second word sense has the synonym set \{recommend, urge, advocate\}\footnote{The word senses and synonyms are from \href{http://wordnetweb.princeton.edu/perl/webwn?s=recommend&sub=Search+WordNet&o2=&o0=1&o8=1&o1=1&o7=&o5=&o9=&o6=&o3=&o4=&h=00000000000000000}{WordNet}.}.
Apparently, the only valid substitution is "commend", which preserves the semantics of the original movie review.
While "urge" is the synonym of "recommend", it obviously does not fit in the context and should not be considered as a possible substitution.
We call substituting $x_i$ with a synonym that matches the word sense of $x_i$ in $\mathbf{x}_{ori}$ a \textit{matched sense substitution}, and we use \textit{mismatched sense substitution} to refer to swapping words with the synonym which belongs to the synonym set of a different word sense.

\subsubsection{Experiments}
\label{subsection:PWWS Experiments}
To illustrate that mismatched sense substitution is a problem existing in practical attack algorithms, we conduct the following analysis.
We examine the adversarial samples generated by PWWS~\citep{ren-etal-2019-generating}, which substitutes words using WordNet synonym set.
We use a benchmark dataset~\citep{yoo-etal-2022-detection} that contains the adversarial samples generated by PWWS against a BERT-based classifier fine-tuned on AG-News~\citep{Zhang2015CharacterlevelCN}.
AG-News is a news topic classification dataset, which aims to classify a piece of news into four categories: world, sports, business, and sci/tech news.
The attack success rate on the testing set composed of 7.6K samples is 57.25\%. 
More statistics about the datasets can be found in Appendix~\ref{appendix:dataset}.
We categorize the words replaced by PWWS into three disjoint categories: \textit{matched sense substitution}, \textit{mismatched sense substitution}, and \textit{morphological substitution}.
The last category, morphological substitution, refers to substituting words with a word that only differs in inflectional morphemes\footnote{Inflectional morphemes are the suffixes that change the grammatical property of a word but do not create a new word, such as a verb's tense or a noun's number. 
For example, recommends$\to$recommend.} or derivational morphemes\footnote{Derivational morphemes are affixes or suffixes that change the form of a word and create a new word, such as changing a verb into a noun form. 
For example, recommend$\to$recommendation.} with the original word.
We specifically isolate \textit{morphological substitution} since it is hard to categorize it into either matched or mismatched sense substitution. 

The detailed procedure of categorizing a replaced word's substitution type is as follows:
Given a pair of $(\mathbf{x}_{ori}, \mathbf{x}_{adv})$, we first use \href{https://www.nltk.org/}{NLTK}~\citep{bird2009natural} to perform word sense disambiguation on each word $x_i$ in $\mathbf{x}_{ori}$.
We use \href{https://github.com/bjascob/LemmInflect}{LemmInflect} and NLTK, to generate the morphological substitution set $\mathbb{ML}_{x_i}$ of $x_i$.
The matched sense substitution set $\mathbb{M}_{x_i}$ is constructed using the WordNet synonym set of the word sense of $x_i$ in $\mathbf{x}_{ori}$; since this synonym set includes the original word $x_i$ and may also include some words in the $\mathbb{ML}_{x_i}$, we remove $x_i$ and words that are already included in the $\mathbb{ML}_{x_i}$ from the synonym set, forming the final matched sense substitution set, $\mathbb{M}_{x_i}$.
The mismatched sense substitution set $\mathbb{MM}_{x_i}$ is constructed by first collecting all synonyms of $x_i$ that belong to the different word sense(s) of $x_i$ in $\mathbf{x}_{ori}$ using WordNet, and then removing all words that have been included in $\mathbb{ML}_{x_i}$ and $\mathbb{M}_{x_i}$.

After inspecting 4140 adversarial samples produced by PWWS, we find that among \textbf{26600} words that are swapped by PWWS, only \textbf{5398 (20.2\%)} words fall in the category of matched sense substitution.
A majority of \textbf{20055 (75.4\%)} word substitutions are mismatched sense substitutions, which should be considered invalid substitutions since using mismatched sense substitution cannot preserve the semantics of $\mathbf{x}_{ori}$ and makes $\mathbf{x}_{adv}$ incomprehensible.
Last, about \textbf{3.8\%} of words are substituted with their morphological related words, such as converting the part of speech (POS) from verb to noun or changing the verb tense.
These substitutions, while maintaining the semantics of the original sentence and perhaps human readable, are mostly ungrammatical and lead to unnatural adversarial samples.
The aforementioned statistics illustrate that only about 20\% word substitutions produced by PWWS are \textit{real} synonym substitutions, and thus the high attack success rate of 57.25\% should not be surprising since most word replacements are highly questionable.

\begin{table*}[t]
    \centering
    \begin{tabular}{|c|ccccc|}
    \hline
    Transformations & Syn. (matched) & Syn. (mismatched) & Antonyms & Morphemes & Others \\
    \hline
    GloVe-kNN & 0.22  & 1.01 & 0 & 1.55 & 27.22\\
    BERT mask-infill & 0.08 & 0.36 & 0.06 & 0.57 & 28.93\\
    BERT reconstruction & 0.14 & 0.58 &  0.09& 1.19 & 27.99\\
    \hline
    \end{tabular}
    \caption{The average words of different substitution types in the candidate word set of $k=$30 words.
    Syn. is short for Synonym.}
    \label{tab:bar.pdf}
\end{table*}

\subsection{Counter-fitted Embedding $k$NN and MLM Mask-Infilling/Reconstruction Contain Few Matched Sense Synonym}
\label{subsection:Counter-fitted Embedding $k$NN and BERT MLM/Reconstruction Contain Few Matched Sense Synonym}
As shown in Section~\ref{subsection:PWWS Experiments}, even when using WordNet synonyms as the candidate sets, the proportion of the valid substitutions is unthinkably low.
This makes us more concerned about the word substitution quality of the other three heuristic transformations introduced in Section~\ref{subsection:Transformation Methods}.
These three word substitution methods mostly rely on assumptions about the quality of the embedding space or the ability of the MLM and require setting a hyperparameter $k$ for the size of the substitution set.
To the best of our knowledge, no previous work has systematically studied what the candidate sets proposed by the three transformations are like; still, they have been widely used in SSAs.

\subsubsection{Experiments}
\label{subsection: BERT Exerperiments}
To understand what those substitution sets are like, we conduct the following experiment.
We use the benchmark dataset generated by~\citet{yoo-etal-2022-detection} that attacks 7.6k samples in the AG-News testing data using TextFooler.
For each word $x_i$ in $\mathbf{x}_{ori}$ that is perturbed into another $x^{'}_{i}$ in $\mathbf{x}_{adv}$, we use the following three transformations to obtain the candidate substitution set: counter-fitted GloVe embedding space, BERT mask-infilling, and BERT reconstruction.
~\footnote{For BERT mask-infilling and reconstruction substitution, we remove punctuation and incomplete subword tokens.
}
We only consider the substitution set of $x_i$ that are perturbed in $\mathbf{x}_{adv}$ because not all words in $\mathbf{x}_{ori}$ will be perturbed by an SSA, and it is thus more reasonable to consider only the words that are really perturbed by an SSA.
We set the $k$ in $k$NN of counter-fitted GloVe embedding space transformation and top-$k$ prediction in BERT mask-infilling/reconstruction to $30$, a reasonable number compared with many previous works.

We categorize the candidate words into five disjoint word substitution types.
Aside from the three word substitution types discussed in Section~\ref{subsection:PWWS Experiments}, we include two other substitution types.
The first one is \textit{antonym substitution}, which is obtained by querying the antonyms of a word $x_i$ using WordNet.
Different from synonym substitutions, we do not separate antonyms into antonyms that matched the word sense of $x_i$ in $\mathbf{x}_{ori}$ and the sense-mismatched antonyms, since neither of them should be considered a valid swap in SSAs.
The other substitution type is \textit{others}, which simply consists of the candidate words not falling in the category of synonyms, antonyms, or morphological substitutions.

In Table~\ref{tab:bar.pdf}, we show how different substitution types comprise the 30 words in the candidate set for different transformations on average.
It is easy to tell that only a slight proportion of the substitution set is made up of synonym substitution for all three transformation methods, with counter-fitted GloVe embedding substitution containing the most synonyms among the three methods, but still only a sprinkle of about 1 word on average.
Moreover, synonym substitution is mostly composed of mismatched sense substitution.
When using BERT mask-infilling as a transformation, there are only 0.08 matched sense substitutions in the top 30 predictions.
While using BERT reconstruction for producing the candidate set, the matched sense substitution slightly increases, compared with mask-infilling, but still only accounts for less than 1 word in the top-30 reconstruction predictions of BERT.
Within the substitution set, there is on average about 1 word which is the morphological substitution of the original word. 
Surprisingly, using MLM mask-infilling or reconstruction as transformation, there is a slight chance that the candidate set consists of antonyms of the original word.
It is highly doubtful whether the semantics is preserved when swapping the original sentence with antonyms.

The vast majority of the substitution set composes of words that do not fall into the previous four categories.
We provide examples of how the substitution sets proposed by different transformations are like in Table~\ref{tab:candidate words} in the Appendix, showing that the candidate words in the \textit{others} substitution types are mostly unrelated words that should not be used for word replacement.
It is understandable that words falling to the \textit{other} substitution types are invalid candidates; this is because the core of SSAs is to replace words with their semantically close counterparts to preserve the semantics of the original sentence.
If a substitution word does not belong to the synonym set proposed by WordNet, it is unlikely that swapping the original word with this word can preserve the semantics of $\mathbf{x}_{ori}$.
We also show some randomly selected adversarial samples generated by different SSAs that use different transformations in Table~\ref{tab:adversarial samples} in the Appendix, which also show that when a word substitution is not a synonym nor a morphological swap, there is a high chance that it is semantically invalid.
~\citet{hauser2021bert} uses human evaluation to show that the adversarial samples generated from TextFooler, BERT-Attack, and BAE do not preserve the meaning of $\mathbf{x}_{ori}$, which also backs up our statement.

When decreasing the number of $k$, the number of invalid substitution words may possibly be reduced.
However, a smaller $k$ often leads to lower attack success rates, as shown in~\citet{li2020bert}, so it is not very common to use a smaller $k$ to ensure the validity of the words in the candidate sets.
In practical attacks, whether these words in the candidate sets can be considered valid depends on the constraints.
But can those constraints really filter invalid substitutions?
We show in the next section that, sadly, the answer is no.

\section{Problems with the Constraints in SSAs}
\label{section:Constraints}

In this section, we show that the constraints commonly used in SSAs cannot fully filter invalid word substitutions proposed by the transformations.

\subsection{Word Embedding Similarity Cannot Distinguish Valid/Invalid Swaps Well}
\label{subsection:Valid Word Substitutions Do Not Necessarily Have Higher Word Embedding Cosine Similarity}
Setting a threshold on word embedding cosine similarity to filter invalid word substitutions relies on the hypothesis that valid word swaps indeed have higher cosine similarity with the word to be substituted, compared with invalid word replacements.
We investigate whether the hypothesis holds with the following experiment.
We reuse the 7.6K AG-News testing samples attacked by TextFooler used in Section~\ref{subsection:Counter-fitted Embedding $k$NN and BERT MLM/Reconstruction Contain Few Matched Sense Synonym}, and we gather all pairs of $(\mathbf{x}_{ori}, \mathbf{x}_{adv})$.
For each word $x_i$ in $\mathbf{x}_{ori}$ that is perturbed in $\mathbf{x}_{adv}$, we follow the same procedure in Section~\ref{subsection:Counter-fitted Embedding $k$NN and BERT MLM/Reconstruction Contain Few Matched Sense Synonym} to obtain the morphological substitution set, matched sense substitution set, mismatched sense substitution set, and the antonym set.
We then query the counter-fitted GloVe embedding space to obtain the word embeddings of all those words and calculate their cosine similarity with the word embedding of $x_i$.
As a random baseline, we also randomly sample high-frequency words and low-frequency words in the training dataset of AG-News, and compute the cosine similarity between those words and $x_i$.
How these high-frequency and low-frequency words are sampled is detailed in Appendix~\ref{appendix:Experiment details for word embedding similarity}.

To quantify how hard it is to use the word embedding cosine similarity to distinguish a valid substitution (the matched sense substitution) from another type of invalid substitution, we calculate the area under the precision-recall curve (AUPR) of the threshold-based detector that predicts whether a perturbed $x^{'}_{i}$ is a valid substitution based on its cosine similarity with $x_i$.
Given an $x_{i}$ and a perturbed $x^{'}_{i}$, a threshold-based detector measures the word embedding cosine similarity between $x_i$ and $x^{'}_{i}$, and assigns it as positive (valid substitution) if the cosine similarity is higher than the threshold.
A perfect detector should have an AUPR of $1.0$, while a random detector will have an AUPR of $0.5$.
Note that the detector we discuss here will only be presented with two types of substitution, one is the matched sense substitution and the other is a substitution type other than the matched sense substitution. 

\begin{table}[t]
    \centering
    \begin{tabular}{|c|c|}
        \hline
        Substitution Type & AUPR \\
        \hline
        Synonyms (mismatched) & 0.627\\
        Antonym & 0.980\\
        Morpheme & 0.433\\
        Random high-freq & 0.900\\
        Random low-freq & 0.919\\
        \hline
    \end{tabular}
    \caption{The AUPR when using a threshold-based detector to separate matched sense synonyms from another type of invalid substitution.}
    \label{tab:AUCROC}
\end{table}

We show the AUPR in Table~\ref{tab:AUCROC}.
First, we notice that when using the word embedding cosine similarity to distinguish matched sense substitutions from mismatched ones, the AUPR is as low as 0.627.
While this is better than random, this is far from a useful detector, showing that word embedding cosine similarity constraints are not useful to remove invalid substitutions like unmatched sense words.
The AUPR for morpheme substitutions is even lower than $0.5$, implying that the word embedding cosine similarity between $x_{i}$ and its morphological similar words is higher than the similarity score between matched sense synonyms.
This means that when we set a higher cosine similarity threshold, we are keeping more morphological swaps instead of valid matched sense substitutions.
While morphological substitutions have meanings similar to or related to the original word, as we previously argued, they are mostly ungrammatical.

The AUPR when using a threshold-based detector to separate matched sense substitutions from antonym substitutions is almost perfect, which is 0.980.
This should not be surprising since the counter-fitted word embedding is designed to make synonyms and antonyms have dissimilar word embeddings.
Last, the AUPR of separating random substitutions from matched sense substitutions is also high, meaning that it is possible to use a detector to remove random and unrelated substitutions based on word embedding cosine similarity.
Based on the result in Table~\ref{tab:AUCROC}, setting a threshold on word-embedding cosine similarity may only filter out the antonyms and random substitutions but still fails to remove the other types of invalid substitutions.

\subsection{Sentence Encoder Is Insensitive to Invalid Word Substitutions}
\label{subsection:Sentence Encoder Is Insensitive to Invalid Word Substitutions}
To test if sentence encoders really can filter invalid word substitutions in SSA, we conduct the following experiment.
We use the same attacked AG-News samples that were used in Section~\ref{subsection: BERT Exerperiments}.
For each pair of $(\mathbf{x}_{ori}, \mathbf{x}_{adv})$ in that dataset, we first collect the swapped indices set $\mathbb{I}=\{i|x_i\neq x_{i}^{'}\}$ that represents the positions of the swapped words in $\mathbf{x}_{adv}$.
We shuffle the elements in $\mathbb{I}$ to form an ordered list $\mathbb{O}$.
Using $\mathbf{x}_{ori}$ and $\mathbb{O}$, we construct a sentence $\mathbf{x}_{swap}^{n}$ by swapping $n$ words in $\mathbf{x}_{ori}$. 
The $n$ positions where the substitutions are made in $\mathbf{x}_{swap}^{n}$ are the first $n$ elements in the ordered list $\mathbb{O}$; at each substitution position, the word is replaced by a word randomly selected from a type of candidate word set.
All the $n$ replaced words in $\mathbf{x}_{swap}^{n}$ are the same type of word substitution.
We conduct experiments with six types of candidate word substitution sets: matched sense, mismatched sense, morphological, antonym, random high-frequency, and random low-frequency word substitutions.
After obtaining $\mathbf{x}_{swap}^{n}$, we compute the cosine similarity between the sentence embedding between $\mathbf{x}_{swap}^{n}$ with $\mathbf{x}_{ori}$ using USE and set the window size $w$ to 7, following~\citet{jin2020bert} and~\citet{garg-ramakrishnan-2020-bae}.
We vary the number of replaced words from 1 to 10.\footnote{Attacking AG-News using TextFooler perturbs about 9 out of 38.6 words in a benign sample on average.}
This experiment helps us know how the cosine similarity changes when the words are swapped using different types of candidate word sets.
More details on this experiment are in Appendix~\ref{appendix:Experiment details for sentence encoder} and Figure~\ref{fig:illustration.pdf} in the Appendix.

\begin{figure}[t]
\centering
\includegraphics[width=0.92\linewidth]{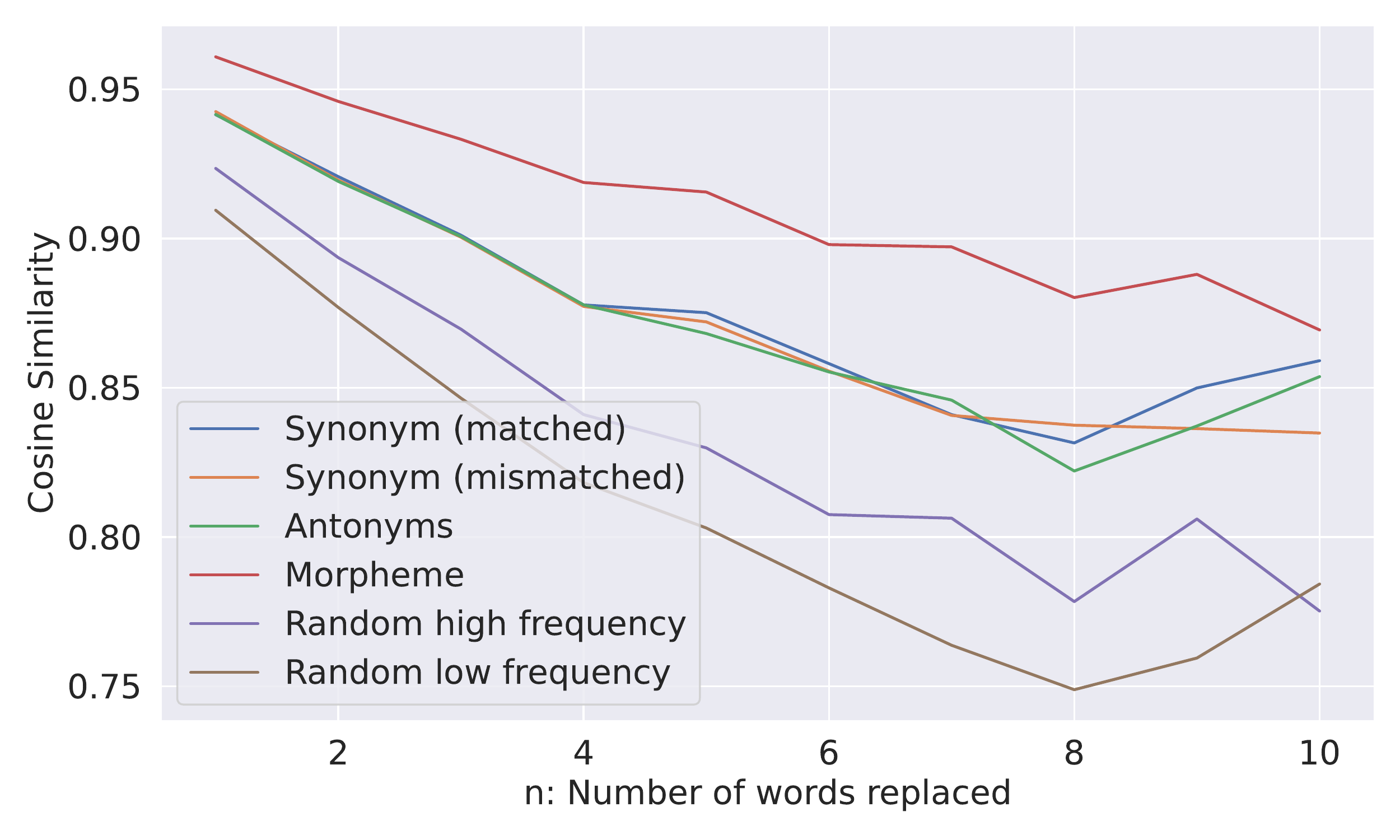}
\caption{
The USE sentence embedding cosine similarity between $\mathbf{x}_{ori}$ and the series of sentences $\mathbf{x}_{swap}^{n}$ obtained by replacing words in $\mathbf{x}_{ori}$ with one type of word substitution.
}
\label{fig:win_original.pdf}
\end{figure}

The results are shown in Figure~\ref{fig:win_original.pdf}.
While replacing more words in $\mathbf{x}_{ori}$ does decrease its cosine similarity with $\mathbf{x}_{ori}$, the cosine similarity when substituting random high-frequency words is still roughly higher than 0.80.
Considering that practical SSAs often set the cosine similarity threshold to around 0.85 or even lower\footnote{We include the sentence embedding cosine similarity threshold of prior works in Table~\ref{tab:SSAs} in Appendix~\ref{appendix: SSAs}.}, depending on the SSAs and datasets, it is suspicious whether the constraint and threshold can really filter invalid word substitution.
We can also observe that when substituting words with antonyms, the sentence embedding cosine similarity with the original sentence closely follows the trend of substituting words using a synonym, regardless of whether the synonym substitution matches the word sense or not.
Recalling that we have revealed that the candidate set proposed by BERT can contain antonyms in Table~\ref{tab:bar.pdf}, the results here indicates that sentence embedding similarity constraint cannot filter this type of faulty word substitution.
For the two different types of synonym substitutions, only matched sense substitutions are valid replacement that follows the semantics of the original sentence.
However, the sentence embedding of $\mathbf{x}_{ori}$ and the sentence embedding of the two types of different synonym substitutions are equally similar.
The highest cosine similarity is obtained when the words in $\mathbf{x}_{ori}$ are swapped using their morphological substitutions, and this is expected since morphological substitutions merely change the semantics. 

In Figure~\ref{fig:win_original.pdf}, we only show the average cosine similarity and do not show the variance of the cosine similarity of each substitution type.
In Figure~\ref{fig:use_variance.pdf} in the Appendix, we show the distribution of the cosine similarity of different substitution types.
The main observation from Figure~\ref{fig:use_variance.pdf} is that the cosine similarity distributions of different substitution types (for the same $n$) are highly overlapped, and it is impossible to distinguish valid word swaps from the invalid ones simply by using a threshold on the sentence embedding cosine similarity.

Overall, the results in Figure~\ref{fig:win_original.pdf} demonstrate that USE tends to generate similar sentence embeddings when two sentences only differ in a few tokens, no matter whether the replacements change the sentence meaning or not.
While we only show the result of USE, we show in Appendix~\ref{appendix:Supplementary Materials for Experiments of Sentence Encoders} that different sentence encoders have similar behavior.
Moreover, when we use the whole sentence instead of a windowed subsentence to calculate the sentence embedding, the cosine similarity is even higher than that shown in Figure~\ref{fig:win_original.pdf}, as shown in Appendix~\ref{appendix:Supplementary Materials for Experiments of Sentence Encoders}.
Again, these sentence encoders fail to separate invalid word substitutions from valid ones.
While frustrating, this result should not be surprising, since most sentence encoders are not trained to distinguish sentences with high word overlapping. 

\subsection{LanguageTool Cannot Detect False Verb Inflectional Form}
\label{subsection:Language-tool Cannot Detect False Verb Inflectional Form}
LanguageTool is used in TextFooler-Adj (TF-Adj)~\citep{morris-etal-2020-reevaluating} to prevent the attack to induce grammar errors.
TF-Adj also uses stricter word embedding and sentence embedding cosine similarity constraints to ensure the semantics in $\mathbf{x}_{ori}$ are preserved in $\mathbf{x}_{adv}$.
However, when browsing through the adversarial samples generated by TF-Adj, we observe that the word substitutions made by TF-Adj are often ungrammatical morphological swaps that convert a verb's inflectional form.
This indicates that LanguageTool may not be capable of detecting a verb's inflectional form error.

To verify this hypothesis, we conduct the following experiment.
For each sample in the test set of AG-News that LanguageTool reports no grammatical errors, we convert the inflectional form of the verbs in the sample by a hand-craft rule that will always make a grammatical sentence ungrammatical; this rule is listed in Appendix~\ref{appendix:language tool experiment details}.
We then use LanguageTool to detect how many grammar errors are there in the verb-converted sentences.

We summarize the experiment results as follows. 
For the 1039 grammatical sentences in AG-News, the previous procedure perturbed \textbf{4.37} verbs on average.
However, the average number of grammar errors identified by LanguageTool is \textbf{0.97}, meaning that LanguageTool cannot detect all incorrect verb forms.
By this simple experiment and the results from Table~\ref{tab:AUCROC} and Figure~\ref{fig:win_original.pdf}, we can understand why the attack results of TF-Adj are often ungrammatical morphological substitutions: higher cosine similarity constraints prefer morphological substitutions, but those often ungrammatical substitutions cannot be detected by LanguageTool.
Thus, aside from showing that the text classifier trained on AG-News is susceptible to inflectional perturbations, TF-Adj actually exposes that LanguageTool itself is vulnerable to inflectional perturbations.

\section{Related Works}
\label{appendix:related works}
Some prior works also discuss a similar question that we study in this paper.
\citet{morris-etal-2020-reevaluating} uses human evaluation to reveal that SSAs sometimes produce low-quality adversarial samples.
They attribute this to the insufficiency of the constraints and use stricter constraints and LanguageTool to generate better adversarial samples.
Our work further points out that the problem is not only in the constraints; we show that the transformations are the fundamental problems in SSAs.
We further show that LanguageTool \citet{morris-etal-2020-reevaluating} uses cannot detect ungrammatical verb inflectional forms, and reveal that the adversarial samples generated by TF-Adj exploit the weakness of LanguageTool and are often made up of ungrammatical morphological substitutions.
~\citet{hauser2021bert} uses human evaluations and probabilistic statements to show that SSAs are low quality and do not preserve original semantics.
Our work can be seen as an attempt to understand the cause of the observations in ~\citet{hauser2021bert}.

~\citet{morris2020second} also questions the validity of using sentence encoders as semantic constraints.
They attack sentence encoders by swapping words in a sentence with their antonyms and the attack goal is to maximally preserve the swapped sentence's sentence embedding cosine similarity with the original sentence. 
This is related to our experiments in Section~\ref{subsection:Sentence Encoder Is Insensitive to Invalid Word Substitutions}.
The main differences between our experiments and theirs are:
(1) When swapping words, we only swap the words that are really swapped by TextFooler; on the contrary, the words swapped in~\citet{morris2020second} are not necessarily words that are actually substituted in an SSA.
The words swapped when attacking a sentence encoder and attacking a text classifier can be significantly different.
Since our goal is to verify how sentence encoders behave when used \textit{in SSAs}, it makes more sense to only swap the words that are really replaced by an SSA.
(2) \citet{morris2020second} only uses antonyms for word substitution.

\section{Discussion and Conclusion}
\label{section:Discussion and Conclusion}
This paper discusses how the elements in SSAs lead to invalid adversarial samples.
We highlight that the candidate word sets generated by all four different word substitution methods contain only a small fraction of semantically matched and grammatically correct word replacements.
While these transformations produce inappropriate candidate words, this alone will not contribute to the invalid adversarial samples.
The inferiority of those adversarial samples should be largely attributed to the deficiency of the constraints that ought to guarantee the quality of the perturbed sentences: word embedding cosine similarity is not always larger for valid word substitutions, sentence encoder is insensitive to invalid word swaps, and LanguageTool fails to detect grammar mistakes.
These altogether bring about the adversarial samples that are human distinguishable, unreasonable, and mostly inexplicable.
These adversarial samples are not suitable for evaluating the vulnerability of NLP models because they are not reasonable inputs. 

The results and observations shown in the main content of our paper are not unique for BERT fine-tuned on AG-News, which is the only attacked model shown in Section~\ref{subsection:Problems with Current Transformation Methods} and Section~\ref{section:Constraints}.
We include supplementary analyses in Appendix~\ref{appendix:Statistics of Other Victim Models and Other Datasets} for different model types and datasets, which supports all the claims and observations in the main contents.
In this paper, we follow previous papers on SSAs to only show the result of attacking the victim model once and not reporting the performance variance due to random seed and hyperparameters used during the fine-tuning of victim model~\citep{ren-etal-2019-generating,li2020bert,jin2020bert}.
This is because conducting SSA is very time-consuming.
In our preliminary experiments, we used TextAttack to attack three BERT models fine-tuned on AG-News and we crafted the adversarial samples for 100 samples in the testing data for each model
The three models were fine-tuned with three different sets of hyperparameters.
We find that our observation in Section~\ref{subsection:Counter-fitted Embedding $k$NN and BERT MLM/Reconstruction Contain Few Matched Sense Synonym} and Section~\ref{section:Constraints} do not change for the three victim models.
Overall, the observation shown in the paper is not an exception but rather a general phenomenon in SSAs.

By the analyses in the paper, we show that we may still be far away from \textit{real} SSAs, and how to construct valid synonym substitution adversarial samples remains an unresolved problem in NLP.
While there is still a long way to go, it is essential to recognize that the prior works have contributed significantly to constructing valid SSAs.
Although prior SSAs may not always produce reasonable adversarial samples, they are still valuable since they pave the way for designing better SSAs and help us uncover the inadequacy of the transformations and constraints for constructing \textit{real} synonym substitution adversarial samples.
As an initiative to stimulate future research, we provide some possible directions and guidelines for constructing better SSAs, based on the observation in our paper.
\begin{enumerate}
    \item Simply consider the word senses when making a replacement with WordNet.
    \item Use better sentence encoders that are sensitive to token replacements that change the semantics of the original sentence. For example, DiffCSE~\citep{chuang2022diffcse} is shown to be able to distinguish the tiny differences between sentences.
    \item When designing transformations, one should always verify the validity of the proposed method through well-controlled experiments. 
    These experiments include recruiting human evaluators to check the quality of the transformations or using experiments as in Section~\ref{subsection:Problems with Current Transformation Methods} to check what the candidate sets proposed by the transformations are like.
    It is perilous to solely rely on heuristics or black-box models such as sentence encoders to guarantee the quality of the transformation.
    \item Since the sentences crafted by SSAs may largely deviate from normal sentences, one should test if constraint models, e.g., grammar checkers or sentence encoders, work as expected when faced with those abnormal sentences. 
    For example, one can perform stress tests~\citep{ribeiro-etal-2020-beyond} to test the behavior of the constraint models.
    This prevents us from exploiting the vulnerability of the constraints when attacking the text classifier.
\end{enumerate}

The problems outlined in this paper may be familiar to those with experience in lexical substitution~\citep{melamud-etal-2015-simple, zhou-etal-2019-bert}, but they have not yet been widely recognized in the field of SSAs. 
Our findings on why SSAs fail can serve as a reality check for the field, which has been hindered by overestimating prior SSAs. 
We hope our work will guide future researchers in cautiously building more effective SSAs.


\section*{Limitations}
In this paper, we only discuss the SSAs in English, as this has been the most predominantly studied in adversarial attacks in NLP.
The authors are not sure whether SSAs in a different language will suffer from the shortcomings discussed in this paper.
However, if an SSA in a non-English language uses the transformations or constraints discussed in this paper, there is a high chance that this attack will produce low-quality results for the same reason shown in this paper.
Still, the above claim needs to be verified by extensive human evaluation and further language-specific analyses.

In our paper, we use WordNet as the gold standard of the word senses since WordNet is a widely adopted and accepted tool in the NLP community. 
Chances are that some annotations in WordNet, while very scarce, are not perfect, and this may be a possible limitation of our work.
It is also possible that the matched sense synonyms found by WordNet may not always be a valid substitution even if the annotation of WordNet is perfect. 
For example, the collocating words of the substituted word may not match that of the original word, and the substitution word may not fit in the original context. 
However, if a word is not even a synonym, it is more unlikely that it is a valid substitution. 
Thus, being a synonym in WordNet is a minimum requirement so we use WordNet synonym sets to evaluate the validity of a word substitution.

Last, we do not conduct human evaluations on what the \textit{other substitution types} in Table~\ref{tab:bar.pdf} are.
As stated in Section~\ref{subsection: BERT Exerperiments}, while we do not perform human evaluations on this, the readers can browse through Table~\ref{tab:candidate words} in the Appendix to see what the \textit{others} substitutions are.
It will be interesting to see what human evaluators will think about the \textit{other} substitutions in the future.

\section*{Ethics Statement and Broader Impacts}
The goal of our paper is to highlight the overlooked details in SSAs that cause their failures.
By mitigating the problems pointed out in our paper, there are two possible consequences:
\begin{enumerate}
    \item One may find that there exist no \textit{real} synonym substitution adversarial samples, and the NLP models currently used are robust. 
    This will cause no ethical concerns since this indicates that no harm will be caused by our work.
    Previous observations on the vulnerability are just the product of low-quality adversarial samples.
    \item There exists \textit{real} synonym substitution adversarial samples, and excluding the issues mentioned in this paper will help malicious users easier to find those adversarial samples. 
    This will become a potential risk in the future.
    The best way to mitigate the above issue is to construct new defenses for \textit{real} SSAs.
\end{enumerate}
While our goal is to raise attention to whether SSAs are really SSAs, we are not advocating malicious users to attack text classifiers using better SSAs.
Instead, we would like to highlight that there is still an unknown risk, the \textit{real} SSAs, against text classifiers, and we researchers should devote more to studying this topic and developing defenses against such attacks before they are adopted by adversarial users.

Another major ethical consideration in our paper is that we challenge prior works on the quality of the SSAs.
While we reveal the shortcomings of previously proposed methods, we still highly acknowledge their contributions.
As emphasized in Section~\ref{section:Discussion and Conclusion}, we do not and try not to devalue those works in the past.
We scientifically and objectively discuss the possible risks of those transformations and constraints, and our ultimate goal is to push the research in adversarial attacks in NLP a step forward; from this perspective, we believe that we are in common with prior works.

\section*{Acknowledgements}
We thank the reviewers for their valuable feedback and actionable suggestions.
We've made major revisions based on the reviews and we list the main modification in Appendix~\ref{App: Different from the Pre-review Version}. 

\bibliography{custom}
\bibliographystyle{acl_natbib}

\appendix

\section{Different from the Pre-review Version}
\label{App: Different from the Pre-review Version}
We list the main difference between this version and the pre-review version of our paper (the pre-review version is similar to the previous arXiv version).
Most modifications are made based on the reviewers' suggestions.
We thank the reviewers for their valuable feedback that help us polish and strengthen this paper.

\begin{itemize}
    \item We change how we present our result in Section~\ref{subsection:Counter-fitted Embedding $k$NN and BERT MLM/Reconstruction Contain Few Matched Sense Synonym} from a bar chart to a table for easier interpretation.
    \item We largely reformulate Section~\ref{subsection:Valid Word Substitutions Do Not Necessarily Have Higher Word Embedding Cosine Similarity}. 
    We change how we present the experiment results: in the previous version, we only qualitatively plot the distribution of the word embedding cosine similarity of different substitution types. 
    In this version, we adopt the reviewers' suggestion to quantitatively show that some types of invalid substitutions cannot be easily detected by the word embedding cosine similarity.
    We also correct the result of antonym substitutions.
    \item We add Section~\ref{appendix:related works} to discuss relevant works.
    \item We discuss the performance variance due to different fine-tuning hyperparameters and random seeds in Section~\ref{section:Discussion and Conclusion}.
    \item We add the links to the victim text classifiers in Appendix~\ref{appendix:dataset}.
    \item We remove the FAQ section in the Appendix, which is mainly used for rebuttal.
    \item In this revision, we incorporate some of the answers to the reviewers' questions in the rebuttal.
\end{itemize}

\section{Dataset}
\label{appendix:dataset}
In our paper, we use benchmark adversarial datasets generated by~\citet{yoo-etal-2022-detection}.
~\citet{yoo-etal-2022-detection} generates adversarial samples using the TextAttack~\citep{morris2020textattack} module.
~\citet{yoo2021towards} release the dataset with a view to facilitating the detection of adversarial samples in NLP and reducing the redundant computation resources to re-generate adversarial samples.
They thus generate adversarial samples using PWWS~\citep{ren-etal-2019-generating}, TextAttack~\citep{jin2020bert}, BAE~\citep{garg-ramakrishnan-2020-bae} and TextFooler-Adj~\citep{morris-etal-2020-reevaluating} on LSTM, CNN, BERT, and RoBERTa trained/fine-tuned on SST-2~\citep{socher-etal-2013-recursive}, IMDB~\citep{maas-etal-2011-learning}, and AG-News~\citep{Zhang2015CharacterlevelCN}.

In the main content of our paper, we only use two datasets: the adversarial samples obtained using PWWS to attack BERT fine-tuned on AG-News, and the adversarial samples obtained by attacking TextFooler on BERT fine-tuned on AG-News.
The testing set of AG-News contains 7.6K samples; the adversarial samples obtained by attacking these datasets will be less than 7.6K since the attack success rates of the two SSAs are not 100\%.
We summarize the detail of these two datasets in Table~\ref{tab:PWWS and TextFooler AG-News}.

The models they used as victim model to generate classifiers are the fine-tuned by the TextAttack~\citep{morris2020textattack} toolkit and are publicly available at \url{https://textattack.readthedocs.io/en/latest/3recipes/models.html} and \href{https://huggingface.co/textattack}{Huggingface models}.
For example, the BERT fine-tuned on AG-News is at \url{https://huggingface.co/textattack/bert-base-uncased-ag-news}.
The hyperparameters used to fine-tune those models can be found from the model cards and \texttt{config.json} and we do not list them here to save the space.

\begin{table*}[th]
    \centering
    \begin{tabular}{|c|c|c|}
    \hline
     & PWWS & TextFooler \\
     \hline
    Success attacks & 4140 & 5885\\
    Attack success rate & 57.25\% & 81.39\% \\
    Average words per sample & 38.57 &38.57 \\
    Average perturbed words percentage & 17.63\% & 23.38\% \\
    \hline
    \end{tabular}
    \caption{Details of the adversarial sample datasets obtained by attacking a BERT fine-tuned on AG-News using PWWS and TextFooler.}
    \label{tab:PWWS and TextFooler AG-News}
\end{table*}

\section{Synonym Substitution Attacks}
\label{appendix: SSAs}
We list the transformations and constraints of the SSAs that are discussed or mentioned in our paper in Table~\ref{tab:SSAs}.
We only include the semantic and grammaticality constraints in Table~\ref{tab:SSAs} and omit other constraints such as the word-level overlap constraints.
The "window" in the sentence encoder cosine similarity constraint indicates whether use a window around the current substitution word or use the whole sentence.
The "compare with $\mathbf{x}_{ori}$" indicates that $\mathbf{x}_{swap}^{n}$ will be compared against the sentence embedding of $\mathbf{x}_{ori}$, and "compared with $\mathbf{x}_{swap}^{n-1}$" means that $\mathbf{x}_{swap}^{n}$ will be compared against the sentence embedding of $\mathbf{x}_{swap}^{n-1}$, that is, the sentence before the current substitution step.

\begin{table*}[]
    \centering
    \begin{tabular}{|p{0.2\linewidth}|p{0.2\linewidth}|p{0.5\linewidth}|}
    \hline
    Attack & Transformation & Constraints \\
    \hline
    Genenetic Algorithm Attack~\citep{alzantot-etal-2018-generating} & Counter-fitted GloVe embedding $k$NN substitution with $k=8$ &  Word embedding mean square error distance with threshold 0.5; language model perplexity (as a grammaticality constraint)\\
    \hline
    PWWS~\citep{ren-etal-2019-generating} &  WordNet synonym set substitution & None \\
    \hline
    TextFooler~\citep{jin2020bert} & Counter-fitted GloVe embedding $k$NN substitution with $k=50$ & USE sentence embedding cosine similarity with threshold 0.878, window size $w=7$, compare with $\mathbf{x}_{ori}$; word embedding cosine similarity with threshold 0.5; disallow swapping words with different POS but allow swapping verbs with nouns or the reverse\\
    \hline
    BERT-Attack~\citep{li2020bert} & BERT mask-infilling substitution with $k=48$ & Sentence embedding cosine similarity with different thresholds for different dataset, and the highest threshold is 0.7, no window, compare with $\mathbf{x}_{ori}$\\
    \hline
    BAE~\citep{garg-ramakrishnan-2020-bae} & BERT reconstruction substitution & USE sentence embedding cosine similarity with threshold 0.936, window size $w=7$, compare with $\mathbf{x}_{swap}^{n-1}$ \\
    \hline
    TextFooler-Adj~\citep{morris-etal-2020-reevaluating} & Counter-fitted GloVe embedding $k$NN substitution with $k=50$ & USE sentence embedding cosine similarity with threshold 0.98, window size $w=7$, compare with $\mathbf{x}_{ori}$; word embedding cosine similarity with threshold 0.9; disallow swapping words with different POS but allow swapping verbs with nouns or the reverse; adversarial sample should not introduce new grammar errors, checked by LanguageTool\\
    \hline
    A2T~\citep{yoo2021towards} & Counter-fitted GloVe embedding $k$NN substitution with $k=20$ or BERT reconstruction with $k=20$ & Word embedding cosine similarity with threshold 0.8; DistilBERT fine-tuned on STS-B sentence embedding cosine similarity with threshold 0.9, window size $w=7$, compare with $\mathbf{x}_{ori}$; disallow swapping words with different POS \\
    \hline
    CLARE~\citep{li-etal-2021-contextualized} & DistilRoBERTa mask-infilling substitution, instead of using top-$k$, they select the predictions whose probability is larger than $5\times 10^{-3}$; this set contains 42 tokens on average & USE sentence embedding cosine similarity with threshold 0.7, window size $w=7$, compare with $\mathbf{x}_{ori}$\\
    \hline
    \end{tabular}
    \caption{Detailed transformations and constraints of different SSAs mentioned in our paper.
    }
    \label{tab:SSAs}
\end{table*}

\subsection{Random Adversarial Samples}
\label{appendix:Random Adversarial Samples}
To illustrate that the adversarial samples generated by SSAs are largely made up of invalid word replacements, we randomly sample two adversarial samples generated by PWWS~\citep{ren-etal-2019-generating}, TextFooler~\citep{jin2020bert}, BAE~\citep{garg-ramakrishnan-2020-bae}, and TextFooler-Adj~\citep{morris-etal-2020-reevaluating}.
To avoid the suspicion of cherry-picking the adversarial samples to support our claims, we simply select the first and the last successfully attacked samples in AG-News using the four SSAs in the dataset generated by~\citet{yoo-etal-2022-detection}.
Since the dataset is not generated by us, we cannot control which sample is the first one and which sample is the last one in the dataset, meaning that we will not be able to cherry-pick the adversarial samples that support our claims.

The adversarial samples are listed in Table~\ref{tab:adversarial samples}.
The \textcolor{blue}{blue} words in $\mathbf{x}_{ori}$ are the words that will be perturbed in $\mathbf{x}_{adv}$.
The \textcolor{red}{red} words are the swapped words.
The readers can verify the claims in our paper using those adversarial samples.
We recap some of our claims as follows:
\begin{itemize}
    \item PWWS uses mismatched sense substitution:
    This can be observed in all the word substitutions of PWWS in Table~\ref{tab:adversarial samples}.
    For example, the word "\textcolor{blue}{world}" in the second example of PWWS have the word sense "the 3rd planet from the sun; the planet we live on".
    But it is swapped with the word "\textcolor{red}{cosmos}", which is the synonym of the word sense "everything that exists anywhere".
    \item Counter-fitted embedding substitution set contains a large proportion of \textit{others} substitution types, which are mostly invalid:
    This can be observed in literally all word substitutions in TextFooler.
    \item BERT reconstruction substitution set contains a large proportion of \textit{others} substitution types, which are mostly invalid:
    This can be observed in literally all word substitutions in BAE.
    \item Morphological substitutions are mostly ungrammatical:
    This can be observed in the first adversarial sample of TextFooler-Adj.
    \item TextFooler-Adj prefers morphological swap due to its strict constraints:
    This can be observe in almost all substitutions in TextFooler-Adj, excluding \textcolor{blue}{goods}$\to$\textcolor{red}{wares}.
\end{itemize}
\begin{table*}[]
    \centering
    \begin{tabular}{|p{0.1\linewidth}|p{0.40\linewidth}|p{0.4\linewidth}|}
    \hline
        Attack & $\mathbf{x}_{ori}$ & $\mathbf{x}_{adv}$ \\
        \hline
        PWWS & 
        Ky. Company \textcolor{blue}{Wins} \textcolor{blue}{Grant} to \textcolor{blue}{Study} Peptides (AP) AP - \textcolor{blue}{A} company founded by a chemistry researcher at the University of Louisville won a grant to develop a method of producing better peptides, which are short chains of amino acids, the building blocks of proteins.
        & 
        Ky. Company \textcolor{red}{profits} \textcolor{red}{yield} to \textcolor{red}{bailiwick} Peptides (AP) AP - \textcolor{red}{amp} company founded by a chemistry researcher at the University of Louisville won a grant to develop a method of producing better peptides, which are short chains of amino acids, the building blocks of proteins. \\
        \hline
        PWWS & 
        Around the \textcolor{blue}{world} Ukrainian presidential candidate Viktor Yushchenko was poisoned with the most harmful known dioxin, which is contained in Agent Orange, a scientist who analyzed his \textcolor{blue}{blood} said Friday.
        & 
        Around the \textcolor{red}{cosmos} Ukrainian presidential candidate Viktor Yushchenko was poisoned with the most harmful known dioxin, which is contained in Agent Orange, a scientist who analyzed his \textcolor{red}{lineage} said Friday. \\
        \hline
        Text- Fooler &
        Fears for T \textcolor{blue}{N} pension after \textcolor{blue}{talks} \textcolor{blue}{Unions} \textcolor{blue}{representing} \textcolor{blue}{workers} at Turner Newall say they are 'disappointed' after \textcolor{blue}{talks} with \textcolor{blue}{stricken} \textcolor{blue}{parent} \textcolor{blue}{firm} \textcolor{blue}{Federal} Mogul.
        & 
        Fears for T \textcolor{red}{percent} pension after \textcolor{red}{debate} \textcolor{red}{Syndicates} \textcolor{red}{portrayal} \textcolor{red}{worker} at Turner Newall say they are 'disappointed' after \textcolor{red}{chatter} with \textcolor{red}{bereaved} \textcolor{red}{parenting} \textcolor{red}{corporations} \textcolor{red}{Canada} Mogul. \\
        \hline
        Text- Fooler & 
        5 of \textcolor{blue}{arthritis} patients in \textcolor{blue}{Singapore} \textcolor{blue}{take} Bextra or Celebrex \&lt; \textcolor{blue}{b}\&gt;...\&\textcolor{blue}{lt};/\textcolor{blue}{b}\&gt; SINGAPORE : \textcolor{blue}{Doctors} in the United \textcolor{blue}{States} \textcolor{blue}{have} \textcolor{blue}{warned} that \textcolor{blue}{painkillers} Bextra and Celebrex may \textcolor{blue}{be} \textcolor{blue}{linked} to \textcolor{blue}{major} cardiovascular \textcolor{blue}{problems} and \textcolor{blue}{should} not be \textcolor{blue}{prescribed}.
        & 
        5 of \textcolor{red}{bursitis} patients in \textcolor{red}{Malaysia} \textcolor{red}{taken} Bextra or Celebrex \&lt;\textcolor{red}{seconds}\&gt;...\&\textcolor{red}{lieutenants};/\textcolor{red}{iii}\&gt; SINGAPORE : \textcolor{red}{Medecine} in the United \textcolor{red}{Nations} \textcolor{red}{get} \textcolor{red}{reminding} that \textcolor{red}{sedatives} Bextra and Celebrex may \textcolor{red}{pose} \textcolor{red}{link} to \textcolor{red}{enormous} cardiovascular \textcolor{red}{woes} and \textcolor{red}{planned} not be \textcolor{red}{planned}. \\
        \hline
        BAE &
        Fears for T \textcolor{blue}{N} pension after talks Unions representing workers at Turner \textcolor{blue}{Newall} say they are 'disappointed' after talks with stricken parent firm Federal Mogul.
        &
        Fears for T \textcolor{red}{pl} pension after talks Unions representing workers at Turner \textcolor{red}{network} say they are 'disappointed' after talks with stricken parent firm Federal Mogul. \\
        \hline
        BAE &
        5 of arthritis patients in Singapore take \textcolor{blue}{Bextra} or \textcolor{blue}{Celebrex} \&\textcolor{blue}{lt};\textcolor{blue}{b}\&gt;...\&lt;/b\&gt; SINGAPORE : \textcolor{blue}{Doctors} in the United \textcolor{blue}{States} have warned that painkillers \textcolor{blue}{Bextra} and Celebrex may be linked to major cardiovascular \textcolor{blue}{problems} and should not be prescribed. 
        &
        5 of arthritis patients in Singapore take \textcolor{red}{cd} or \textcolor{red}{i} \&\textcolor{red}{m};\textcolor{red}{x}\&gt;...\&lt;/b\&gt; SINGAPORE : \textcolor{red}{doctors} in the United \textcolor{red}{state} have warned that painkillers \textcolor{red}{used} and Celebrex may be linked to major cardiovascular \textcolor{red}{harm} and should not be prescribed. \\
        \hline
        Text- Fooler -Adj &
        Venezuela Prepares for Chavez Recall \textcolor{blue}{Vote} Supporters and rivals warn of possible fraud; government says Chavez's defeat could produce turmoil in world oil \textcolor{blue}{market}.
        &
        Venezuela Prepares for Chavez Recall \textcolor{red}{Voted} Supporters and rivals warn of possible fraud; government says Chavez's defeat could produce turmoil in world oil \textcolor{red}{marketed}. \\
        \hline
        Text- Fooler -Adj &
        EU to Lift U.S. Sanctions Jan. 1  BRUSSELS (Reuters) - The European Commission is sticking with its plan to lift sanctions on \$4 billion worth of U.S. \textcolor{blue}{goods} on Jan. 1 following Washington's repeal of export \textcolor{blue}{tax} subsidies in October, a spokeswoman said on Thursday.
        &
        EU to Lift U.S. Sanctions Jan. 1  BRUSSELS (Reuters) - The European Commission is sticking with its plan to lift sanctions on \$4 billion worth of U.S.  \textcolor{red}{wares} on Jan. 1 following Washington's repeal of export \textcolor{red}{taxation}  subsidies in October, a spokeswoman said on Thursday. \\
        \hline
        
    \end{tabular}
    \caption{Adversarial samples from the benchmark dataset generated by~\citet{yoo2021towards}.}
    \label{tab:adversarial samples}
\end{table*}

\subsubsection{Example of the Word Substitution Sets of Different Transformations}
\label{appendix:Example of the Word Substitution Sets of Different Transformations}
In this section, we show the substitution sets using different transformations.
We only show one example here, and this example is the second successful attack example in the adversarial sample datasets~\citep{yoo-etal-2022-detection} that attacks a BERT fine-tuned classifier trained on AG-News using TextFooler.
We do not use the first sample in Table~\ref{tab:adversarial samples} because we would like to show the readers a different adversarial sample in the datasets.

$\mathbf{x}_{ori}$: The Race is \textcolor{blue}{On}: Second \textcolor{blue}{Private} \textcolor{blue}{Team} \textcolor{blue}{Sets} \textcolor{blue}{Launch} \textcolor{blue}{Date} for \textcolor{blue}{Human} Spaceflight (\textcolor{blue}{SPACE}.com) \textcolor{blue}{SPACE}.com - \textcolor{blue}{TORONTO}, \textcolor{blue}{Canada} -- \textcolor{blue}{A} second \textcolor{blue}{team} of rocketeers \textcolor{blue}{competing} for the  \#36;10 \textcolor{blue}{million} Ansari X \textcolor{blue}{Prize}, a \textcolor{blue}{contest} for \textcolor{blue}{privately} \textcolor{blue}{funded} suborbital \textcolor{blue}{space} \textcolor{blue}{flight}, \textcolor{blue}{has} \textcolor{blue}{officially} \textcolor{blue}{announced} the first \textcolor{blue}{launch} \textcolor{blue}{date} for its \textcolor{blue}{manned} \textcolor{blue}{rocket}.

$\mathbf{x}_{adv}$: The Race is \textcolor{red}{Around}: Second \textcolor{red}{Privy} \textcolor{red}{Remit} \textcolor{red}{Set} \textcolor{red}{Lanza} \textcolor{red}{Timeline} for \textcolor{red}{Hummanitarian} Spaceflight (\textcolor{red}{SEPARATION}.com) \textcolor{red}{SEPARATION}.com - \textcolor{red}{CANADIENS}, \textcolor{red}{Countries} -- \textcolor{red}{para} second \textcolor{red}{squad} of rocketeers \textcolor{red}{suitors} for the  \#36;10 \textcolor{red}{billion} Ansari X \textcolor{red}{Nobel}, a \textcolor{red}{contestant} for \textcolor{red}{convertly} \textcolor{red}{championed} suborbital \textcolor{red}{spaceship} \textcolor{red}{plane}, \textcolor{red}{had} \textcolor{red}{solemnly} \textcolor{red}{proclaim} the first \textcolor{red}{began} \textcolor{red}{timeline} for its \textcolor{red}{desolate} \textcolor{red}{bomb}.

We show the substitution set for the first four words that are substituted by TextFooler in Table~\ref{tab:candidate words}.
We do not show that substitution set for all the attacked words simply because it will occupy too much space, and our claim in the main content that "\textit{others} substitution sets of counter-fitted embedding substitution and BERT mask-infilling/reconstruction mostly consist of invalid swaps" can already be observed in Table~\ref{tab:candidate words}.

\begin{table*}[]
    \centering
    \begin{tabular}{|p{0.06\linewidth}|p{0.28\linewidth}|p{0.22\linewidth}|p{0.25\linewidth}|}
    \hline
    $x_i$ & Counter-fitter GloVe embedding & BERT MLM & BERT reconstruction \\
    \hline
    On
    &
    Orn, Pertaining, Per, Toward, Dated, Towards, Circa, Dates, Relating, Pour, Relative, Sur, Into, Date, Concerning, Onto, Around, About, In, To, Sobre, Relate, During, Respecting, For, Regarding, At, Days, Throughout, Relation
    & 
    following, completed, ongoing, over, in, included, contested, followed, this, now, below, announced, after, split, for, therefore, concluded, titled, currently, follows, planned, listed, thus, held, on, to, that, scheduled, called, where
    &
    around, round, a, here, ongoing, over, in, the, involved, pending, at, next, now, under, for, ahead, set, \textcolor{purple}{off}, currently, onto, given, considered, about, held, on, of, to, by, time, with\\
    \hline
    Private
    &
    Confidentiality, Camera, Personal, Clandestine, Privately, Hoc, Undercover, Confidential, Secretive, Secrets, Dedicated, \textcolor{orange}{Secret}, Surreptitiously, Confidentially, Belonged, Peculiar, Personally, Specially, Fenced, Owned, Covert, Particular, Especial, Covertly, Own, Deprived, Secretly, Privy, Soldier, Special
    &
    google, my, o, a, from, hs, the, 1, chapter, 1st, in, this, mv, md, ukrainian, le, facebook, baltimore, hr, of, th, to, that, donald, and, by, gma, where, with
    &
    personal, vr, 2012, my, a, from, own, official, local, the, vc, small, for, national, billionaire, social, private, 2014, 2010, pv, facebook, \textcolor{purple}{public}, independent, of, privately, to, new, family, and, by
    \\
    \hline
    Team &
    Panels, Grouping, Machine, Equipments, Tasks, Task, Devices, Pc, Group, Appliance, Cluster, Computers, Groups, \textcolor{brown}{Teams}, Tooling, Accoutrements, Remit, Pcs, Appliances, Grupo, Teamwork, Chore, Apparatus, \textcolor{orange}{Squad}, Computer, Device, Machines, Panel, Squads, Equipment
    &
    fund, label, launch, google, team, sponsor, investor, project, citizen, investigator, sector, plane, foundation, company, helicopter, website, line, platform, rocket, and, group, blog, planet, computer, charity, to, jet, pilot, party, fan
    &
    firm, one, weekend, partnership, round, \textcolor{brown}{team}, committee, \textcolor{brown}{teams}, number, couple, country, site, button, company, line, side, crew, ballot, group, nation, winner, division, club, boat, of, to, family, party, time
    \\
    \hline
    Sets & 
    Defines, Stake, Matches, Provides, Prescribes, \textcolor{orange}{Determine}, \textcolor{brown}{Set}, Betting, Establishes, Stipulates, Jeu, Gambling, Staking, Stipulated, Toys, Determines, Defined, Game, Defining, Playing, Gaming, Games, Determining, \textcolor{orange}{Define}, Jeux, Gamble, Identifies, Stipulate, Plays, Play
    &
    google, a, from, estimated, first, larsen, the, 1, 1st, 3, at, next, announced, top, named, def, or, possible, predicted, 3rd, facebook, 000, online, about, on, of, to, and, no, with
    &
    reaches, established, announce, places, records, official, announcing, begins, forms, indicates, announced, declares, sets, starts, estimates, \textcolor{orange}{determines}, \textcolor{brown}{set}, details, draws, lays, lists, specifies, calls, \textcolor{brown}{setting}, stages, of, gives, establishes, announces, names
    \\
    \hline
    \end{tabular}
    \caption{Candidate substitutions proposed by different transformations. 
    We use \textcolor{green}{green} to denote matched sense substitution, \textcolor{orange}{orange} to denote mismatched sense substitution, \textcolor{brown}{brown} to denote morpheme substitution, and \textcolor{purple}{purple} to denote antonyms.
    The \textit{other} type substitution is denoted using the default black.}
    \label{tab:candidate words}
\end{table*}

\section{Implementation Details}
\label{appendix:Implementation Details}

\subsection{Experiment Details of Section~\ref{subsection:Problems with Current Transformation Methods}}
\label{appendix:Experiment details for word substitution}
In this section, we give details on how we obtain different word substitution types for a $\mathbf{x}_{ori}$.
The whole process is summarized in Algorithm~\ref{alg:transformation}.
In Algorithm~\ref{alg:transformation}, the reader can also find how the perturbed indices list $\mathbb{I}$ used in Section~\ref{subsection:Sentence Encoder Is Insensitive to Invalid Word Substitutions} is obtained.

An important detail that is not mentioned in the main content is that when computing how many synonyms are in the substitution set of BERT MLM substitution, we actually perform lemmatization on the top-30 predictions of BERT.
This is because, for example, if BERT proposes to use the word "defines" to replace the original word "sets" (the third person present tense of the verb "set") in the original sentence, and the word "define" happens to a synonym according to WordNet; in this case, the word "defines" will not be considered as a synonym substitution.
But "defines" should be considered as a synonym substitution since it is the third person present tense of "define".
Lemmatizing the prediction of BERT can partially solve the problem.
However, if the lemmatized word is already in the top-30 prediction of BERT, we do not perform lemmatization.
This process is detailed on Line~\ref{alg:mlm:if} on Algorithm~\ref{alg:mlm}.
This can ensure that words can be considered as synonyms while words that should be considered as morphological swaps are mostly not affected.

\begin{algorithm*}
\caption{Process of obtaining the substitution set}
\label{alg:transformation}
\begin{algorithmic}[1]
\Require $\mathbf{x}_{ori}, \mathbf{x}_{adv}$
\State $\mathbb{I} \gets []$ \Comment{Initialize the perturbed indices list}
\For {$x_i \in \mathbf{x}_{ori}$} 
    \If{$x_i = x_{i}^{'}$} 
        \State \textbf{continue}
    \EndIf
    \State $x_i \gets x_i.\text{lower()}$ \Comment{Get the lower case of $x_i$}
    \State $x_i^{'} \gets x_i^{'}.\text{lower()}$ \Comment{Get the lower case of $x_i^{'}$}
    \State $\mathbb{S}_{ml} \gets \textbf{GetMorph}(x_i, \mathbf{x}_{ori})$ 
    \Comment{Get morphological substitutions}
    \State $\mathbb{S}_{ms} \gets \textbf{GetMatchedSense}(x_i, \mathbf{x}_{ori})$
    \Comment{Get matched sense synonym by first using word sense disambiguation then WordNet synonym sets}
    \State $\mathbb{S}_{mms} \gets \textbf{GetMismatchedSense}(x_i, \mathbf{x}_{ori})$
    \Comment{Get mismatched sense synonym by first using word sense disambiguation then WordNet synonym sets}
    \State $\mathbb{A} \gets \textbf{GetAntonym}(x_i)$
    \Comment{Get antonyms by WordNet}
    
    \State $\mathbb{S}_{ml}\gets \mathbb{S}_{ml}\setminus\{x_i\}$ 
    \State $\mathbb{S}_{ms}\gets \mathbb{S}_{ms}\setminus\mathbb{S}_{ml}\setminus\{x_i\}$
    \State $\mathbb{S}_{mms}\gets \mathbb{S}_{mms} \setminus \mathbb{S}_{ms} \setminus \mathbb{S}_{ml} \setminus \{x_i\}$
    \Comment{Remove overlapping elements to make $\mathbb{S}_{ml}$, $\mathbb{S}_{ms}$, $\mathbb{S}_{mms}$ disjoint}

    \State $\mathbb{S}_{embed} \gets \textbf{GetEmbeddingSwaps}(x_i)$ 
    \State $\mathbb{S}_{MLM} \gets \textbf{GetMLMSwaps}(x_i,\mathbf{x}_{ori})$ 
    \State $\mathbb{S}_{recons} \gets \textbf{GetReconsSwaps}(x_i,\mathbf{x}_{ori})$ 
    
    \If{$x_i^{'}\in \mathbb{S}_{ml} $}
      \State\text{The substitution is a morphological substitution}
    \ElsIf{$x_i^{'}\in \mathbb{S}_{ms} $}
      \State\text{The substitution is a matched sense substitution}
    \ElsIf{$x_i^{'}\in \mathbb{S}_{mms} $}
      \State\text{The substitution is a mismatched sense substitution}
    \ElsIf{$x_i^{'}\in \mathbb{A} $}
      \State\text{The substitution is an antonym substitution}
    \Else
      \State \text{This substitution is a \textit{other} type}
    \EndIf
    
    \State {Check the substitution types of each word in $\mathbb{S}_{embed}$ by comparing with  $\mathbb{S}_{ml}$, $\mathbb{S}_{ms}$, $\mathbb{S}_{mms}, \mathbb{A}$ } 
    \State {Check the substitution types of each word in $\mathbb{S}_{MLM}$ by comparing with  $\mathbb{S}_{ml}$, $\mathbb{S}_{ms}$, $\mathbb{S}_{mms}, \mathbb{A}$} 
    \State {Check the substitution types of each word in $\mathbb{S}_{recons}$ by comparing with  $\mathbb{S}_{ml}$, $\mathbb{S}_{ms}$, $\mathbb{S}_{mms}, \mathbb{A}$} 
    \If{$\mathbb{S}_{ml},\mathbb{S}_{ms}, \mathbb{S}_{mms}, \mathbb{A}\neq \emptyset$}
        \State $\mathbb{I}$\text{.append}($i$) 
        \Comment{We only include the words whose have morphological substitutions, matched sense substitutions, mismatched sense substitutions}
    \EndIf
\EndFor

\State $\mathbb{O}\gets$\text{shuffle}$.(\mathbb{I})$
\end{algorithmic}
\end{algorithm*}

\begin{algorithm*}
\caption{\textbf{GetMLMSwaps}$x_i,\mathbf{x}_{ori}$}
\label{alg:mlm}
\begin{algorithmic}[1]
\Require $x_i, \mathbf{x}_{ori}$, BERT, Lemmatizer
\State $\mathbf{x}_{mask} \gets \{x_1, \cdots, x_{i-1}, \text{[MASK]}, x_{i+1},\cdots,x_n\}$ \Comment{Get masked input}
\State Candidates$\gets$ Top-k prediction of $\mathbf{x}_{mask}$ using BERT
\State New\_Candidates $\gets []$ 
\For{$w\in$Candidates}
    \State $w_{lemmatized}\gets$ Lemmatizer$(w)$
    \If{$w_{lemmatized}\notin$Candidates and $w_{lemmatized}\notin$New\_Candidates} \label{alg:mlm:if}
        \State New\_Candidates.append($w_{lemmatized}$)
    \Else
        \State New\_Candidates.append($w$)
    \EndIf
\EndFor
\State \Return New\_Candidates
\end{algorithmic}
\end{algorithm*}

\subsection{Experiment Details of Section~\ref{subsection:Valid Word Substitutions Do Not Necessarily Have Higher Word Embedding Cosine Similarity}}
\label{appendix:Experiment details for word embedding similarity}
Here, we explain how the random high/low-frequency words are sampled in Section~\ref{subsection:Valid Word Substitutions Do Not Necessarily Have Higher Word Embedding Cosine Similarity}.
First, we use the tokenizer of BERT-base-uncased to tokenize all the samples in the training dataset of AG-News.
Next, we count the occurrence of each token in the vocabulary of the BERT-base-uncased, and sort the tokens based on their occurrence in the training set in descending order.
The vocabulary size of BERT-base-uncased is 30522, including five special tokens, some subword tokens, and some unused tokens.
We define the high-frequency words as the top-50 to top-550 words in the training dataset.
The reason that we omit the top 50 words as the high-frequency token is that these words are often stop words, and they are seldom used as word substitutions in SSAs.
The low-frequency words are the top-10K to top-10.5K occurring words in AG-News' training set.

\subsection{Experiment Details of Section~\ref{subsection:Sentence Encoder Is Insensitive to Invalid Word Substitutions}}
\label{appendix:Experiment details for sentence encoder}
Here, we give more details on the sentence embedding similarity experiment in Section~\ref{subsection:Sentence Encoder Is Insensitive to Invalid Word Substitutions}.
The readers can refer to Algorithm~\ref{alg:transformation} to see how we obtain the different types of word substitution sets, the substituted indices set $\mathbb{I}$ and the ordered list $\mathbb{O}$ from a pair of $(\mathbf{x}_{ori},\mathbf{x}_{adv})$.

\begin{figure*}[t]
\centering
\includegraphics[width=1.0\linewidth]{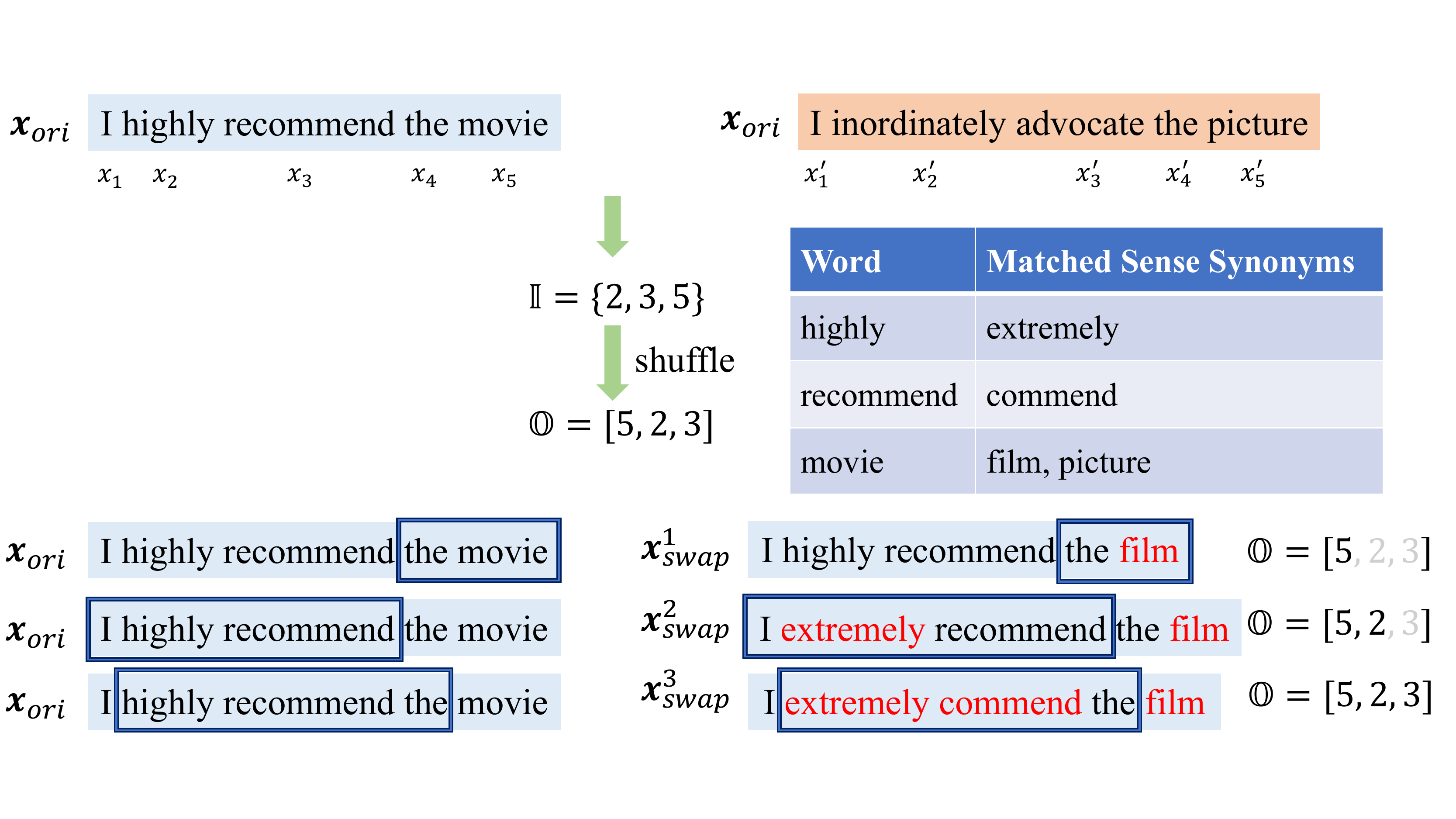}
\caption{ 
An example for illustrating process of obtaining $\mathbb{I},\mathbb{O}$ and $\mathbf{x}^{n}_{swap}$ from a pair of $(\mathbf{x}_{ori},\mathbf{x}_{adv})$.
Here, the substitution type used for constructing $\mathbf{x}^{n}_{swap}$ is the matched sense synonyms. 
The subsentences outlined by dark blue in the bottom three $\mathbf{x}_{ori}$ and $\mathbf{x}_{swap}^{n}$ are the parts that are used to compute the sentence embedding by the sentence encoder.
In the figure, we set the window size $w$ of the sentence encoder to $1$ for the ease of illustration.
}
\label{fig:illustration.pdf}
\end{figure*}

We also use a figurative illustration to show how we obtain $\mathbf{x}_{swap}^{n}$ in Figure~\ref{fig:illustration.pdf}.
In Figure~\ref{fig:illustration.pdf}, we show how to use the \textit{same sense substitution set} to replace the words in $\mathbf{x}_{ori}$ based on the ordered list $\mathbb{O}$.
As can be seen in the figure, we swap the words in $\mathbf{x}_{ori}$ according to the order determined by $\mathbb{O}$; since the first element in $\mathbb{O}$ is $5$, we will first replace $x_5$ in $\mathbf{x}_{ori}$ with one of the same sense synonyms of $x_5$.
We thus obtain the $\mathbf{x}_{swap}^{1}$.
In order to compute the sentence embedding similarity between $\mathbf{x}_{swap}^{1}$ and $\mathbf{x}_{ori}$, we extract a context around the word just replaced; in this case, we will extract the context around the fifth word in $\mathbf{x}_{swap}^{1}$ and $\mathbf{x}_{ori}$.
Different from what we really use in our experiment, we set the window size $w$ to 1 in Figure~\ref{fig:illustration.pdf}; this is because using $w=7$ is too large for this example.
Thus, we should extract $\mathbf{x}_{swap}^{1}[4:7]$ and  $\mathbf{x}_{ori}[4:7]$; however, since the sentences only have 5 words, the context to be extracted will exceed the length of the sentences.
In this case, we simply extract the context until the end of both sentences.\footnote{Similarly, if the context to be extracted starts from a position that is on the left-hand side of the sentence, we simply extract the context starting from the first word in the sentence.}
The parts that will be used for computing the sentence embeddings in each sentence are outlined with a dark blue box in Figure~\ref{fig:illustration.pdf}.
Next, we follow a similar process to obtain $\mathbf{x}_{swap}^{2}$ and $\mathbf{x}_{swap}^{3}$ and compare their sentence embedding cosine similarity with $\mathbf{x}_{ori}$.

\subsection{Experiment Details of Section~\ref{subsection:Language-tool Cannot Detect False Verb Inflectional Form}}
\label{appendix:language tool experiment details}
In this experiment, we usethe POS tagger in NLTK to identify the verb form of the verbs.
The inflectional form of the verbs are obtained using LemmInflect.
Here, we list the verb inflectional form conversion rules:
\begin{itemize}
    \item For each third-person singular present verb, it is converted to the verb's base form.
    \item For each third past tense verb, it is converted to the verb's gerund or present participle form (V+ing).
    \item For all verbs whose form is not third-person singular present and is not past tense verb, we convert them into the third-person singular present.
We provide three random examples from the test set in AG-News that are perturbed using the above rules in Table~\ref{tab:verb-perturbed examples}.
It can be easily seen that all the perturbed sentences are ungrammatical.
Interestingly, LanguageTool detects no grammar errors in all the six samples in Table~\ref{tab:verb-perturbed examples}.

\begin{table*}[]
    \centering
    \begin{tabular}{|p{0.45\linewidth}|p{0.45\linewidth}|}
    \hline
    Original sentence & Verb-perturbed sentence \\ \hline
    Storage, servers bruise HP earnings update Earnings per share rise \textcolor{blue}{compared} with a year ago, but company \textcolor{blue}{misses} analysts' expectations by a long shot. & 
    Storage, servers bruises HP earnings update Earnings per share rise \textcolor{red}{compares} with a year ago, but company \textcolor{red}{miss} analysts' expectations by a long shot. \\ \hline
    IBM to \textcolor{blue}{hire} even more new workers By the end of the year, the computing giant plans to \textcolor{blue}{have} its biggest headcount since 1991. &
    IBM to \textcolor{red}{hires} even more new workers By the end of the year, the computes giant plans to \textcolor{red}{has} its biggest headcount since 1991. \\ \hline
    Giddy Phelps Touches Gold for First Time Michael Phelps \textcolor{blue}{won} the gold medal in the 400 individual medley and \textcolor{blue}{set} a world record in a time of 4 minutes 8.26 seconds. & 
    Giddy Phelps Touches Gold for First Time Michael Phelps \textcolor{red}{winning} the gold medal in the 400 individual medley and \textcolor{red}{sets} a world record in a time of 4 minutes 8.26 seconds. \\ \hline
        
    \end{tabular}
    \caption{Examples of the verb-perturbed sentences. The perturbed verbs are highlighted in red, and their unperturbed counterparts are highlighted in blue.}
    \label{tab:verb-perturbed examples}
\end{table*}

\end{itemize}

\section{Supplementary Materials for Experiments of Sentence Encoders}
\label{appendix:Supplementary Materials for Experiments of Sentence Encoders}
\subsection{Distribution of the Sentence Embedding Cosine Similarity of Different Substitution Types}
\begin{figure*}[t]
\centering
\includegraphics[width=1.0\linewidth]{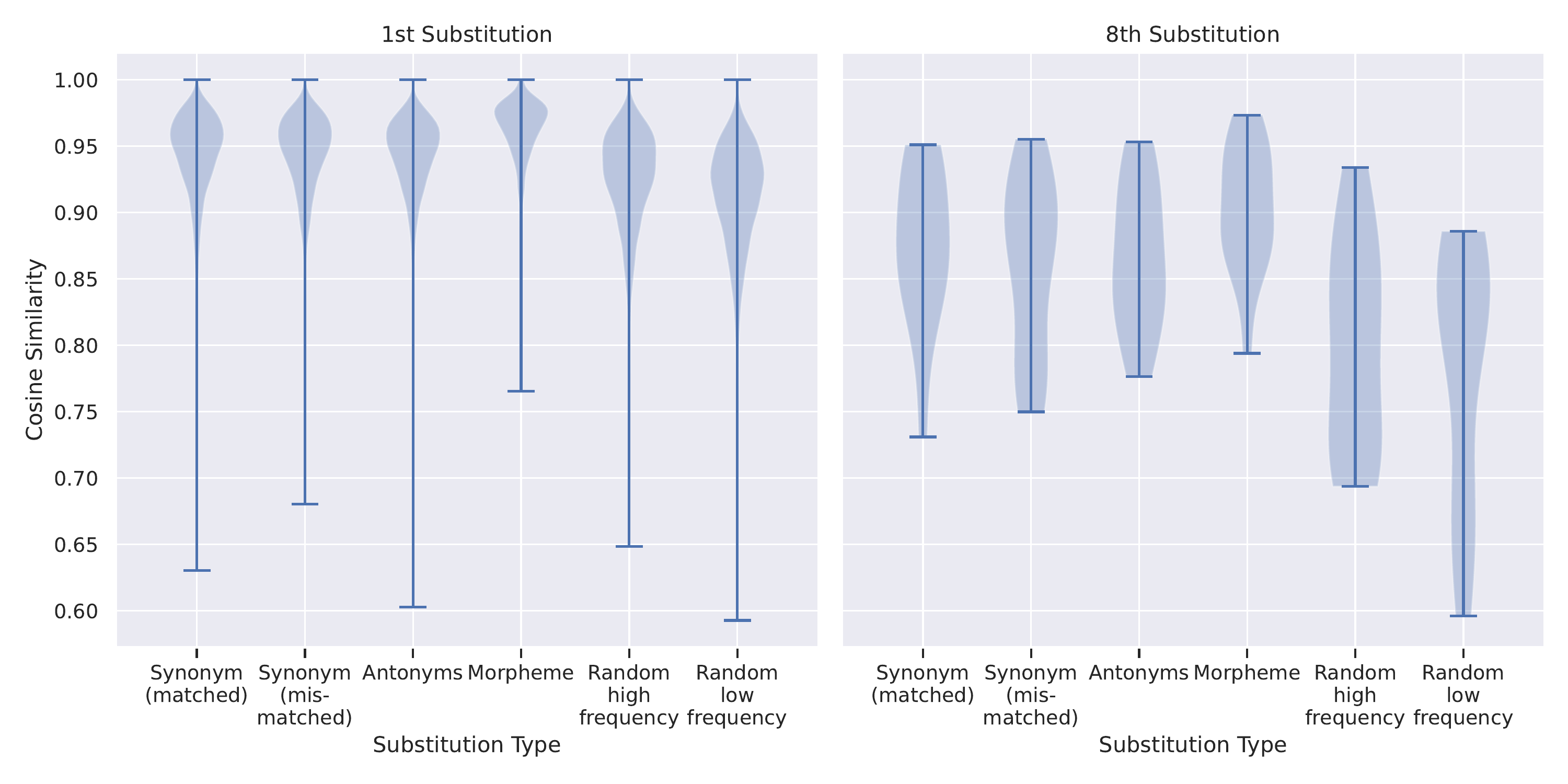}
\caption{ 
The USE sentence embedding cosine similarity distribution between $\mathbf{x}_{ori}$ and the series of sentences obtained by replacing words in $\mathbf{x}_{ori}$ with one type of word substitution.
The window size is the same as Figure~\ref{fig:win_original.pdf}.
The left subfigure shows the distribution of the cosine similarity between $\mathbf{x}_{ori}$ and $\mathbf{x}_{swap}^{1}$ and the right subfigure is the similarity distribution between $\mathbf{x}_{ori}$ and $\mathbf{x}_{swap}^{8}$.
}
\label{fig:use_variance.pdf}
\end{figure*}

In Figure~\ref{fig:use_variance.pdf}, we show the distribution of the USE sentence embedding cosine similarity of different word replacement types using different numbers of word replacements $n$.
The left subfigure shows the distribution of the cosine similarity between $\mathbf{x}_{ori}$ and $\mathbf{x}_{swap}^{1}$ and the right subfigure is the similarity distribution between $\mathbf{x}_{ori}$ and $\mathbf{x}_{swap}^{8}$.
While in Figure~\ref{fig:win_original.pdf}, we can see that the sentence embedding cosine similarity of different word substitution types is sometimes separable on average, we still cannot separate valid and invalid word substitution simply using one threshold.
This is because the word embedding cosine similarity scores of different word substitution types are highly overlapped, which is evident from Figure~\ref{fig:use_variance.pdf}.
This is true for different $n$ of $\mathbf{x}_{swap}^{n}$, and we only show $n=1$ and $n=8$ for simplicity.

\subsection{Different Methods For Computing Sentence Embedding Similarity}

\begin{figure}[t]
\centering
\includegraphics[width=1.0\linewidth]{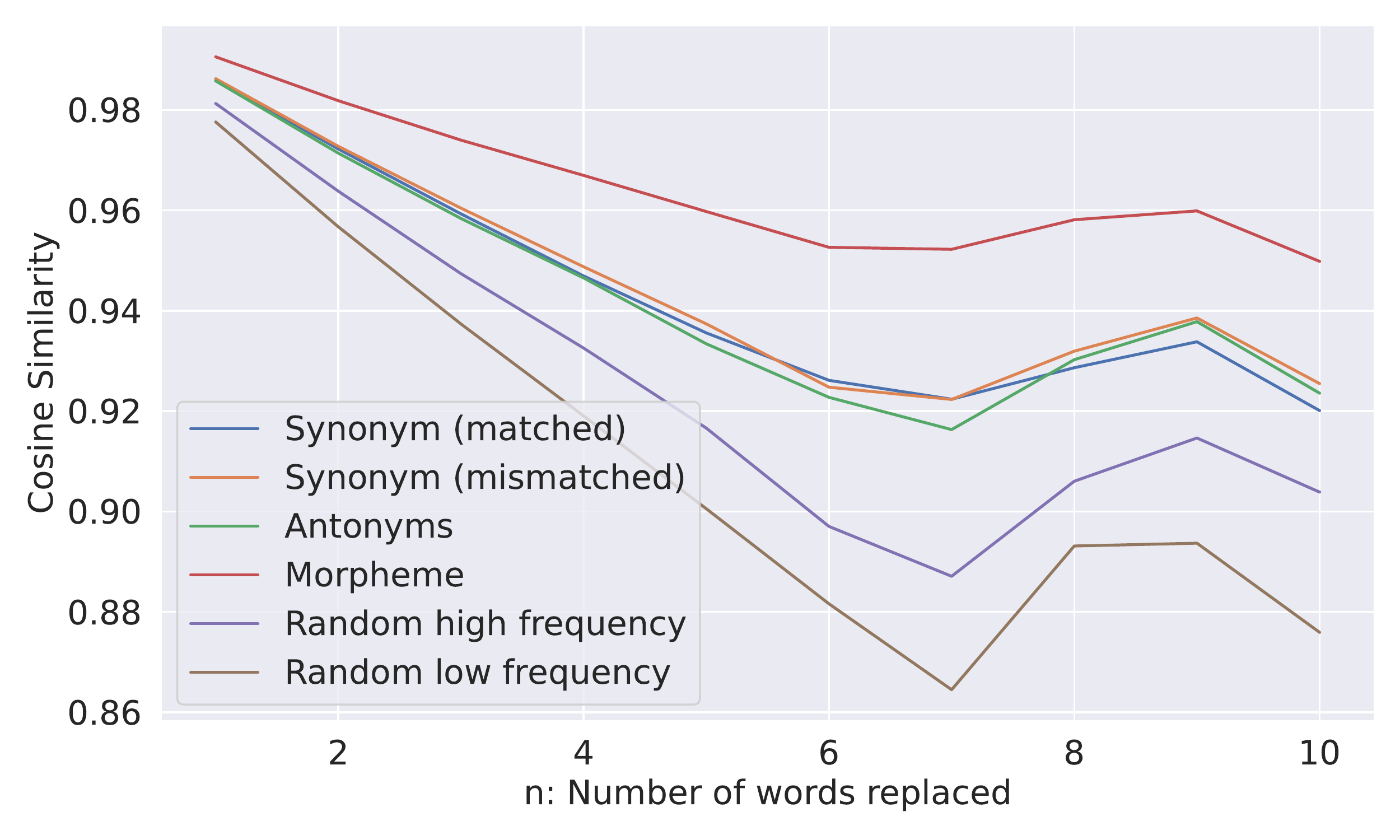}
\caption{ 
The USE sentence embedding cosine similarity between $\mathbf{x}_{ori}$ and the series of sentences obtained by replacing words in $\mathbf{x}_{ori}$ with one type of word substitution.
Different from Figure~\ref{fig:win_original.pdf}, we use the whole sentence (without using window) to compute the sentence embedding of $\mathbf{x}_{ori}$ and $\mathbf{x}_{swap}^{n}$.
}
\label{fig:no_win_original.pdf}
\end{figure}

In this section, we show some supplementary figures of the experiments in Section~\ref{subsection:Sentence Encoder Is Insensitive to Invalid Word Substitutions}.
Recall that in the main content, we only show the sentence embedding cosine similarity results when we compare $\mathbf{x}_{swap}^{n}$ with $\mathbf{x}_{ori}$ around a 15-word window around the $n$-th substituted word.
But we have mentioned in Section~\ref{subsection:Constraints} that this is not what is always done.
In Figure~\ref{fig:no_win_original.pdf}, we show the result when we compare $\mathbf{x}_{swap}^{n}$ with $\mathbf{x}_{ori}$ using \textbf{the whole sentence}.
It can be easily observed that it is still difficult to separate valid swaps from the invalid ones using a threshold on the cosine similarity.
One can also observe that the similarity in Figure~\ref{fig:no_win_original.pdf} is a lot higher than that in Figure~\ref{fig:win_original.pdf}.

Another important implementation detail about sentence encoder similarity constraint is that some previous work does not calculate the similarity of the current $\mathbf{x}_{swap}$ with $\mathbf{x}_{ori}$.
Instead, they calculate the similarity between the current $\mathbf{x}_{swap}$ and the $\mathbf{x}_{swap}$ in the previous substitution step~\citep{garg-ramakrishnan-2020-bae}.
That is, if in the previous substitution step, 6 words in $\mathbf{x}_{ori}$ are swapped, and in this substitution step, we are going to make the 7th substitution. 
Then the sentence embedding similarity is computed between the 6-word substituted sentence and the 7-word substituted sentence.

\begin{figure}[t]
\centering
\includegraphics[width=1.0\linewidth]{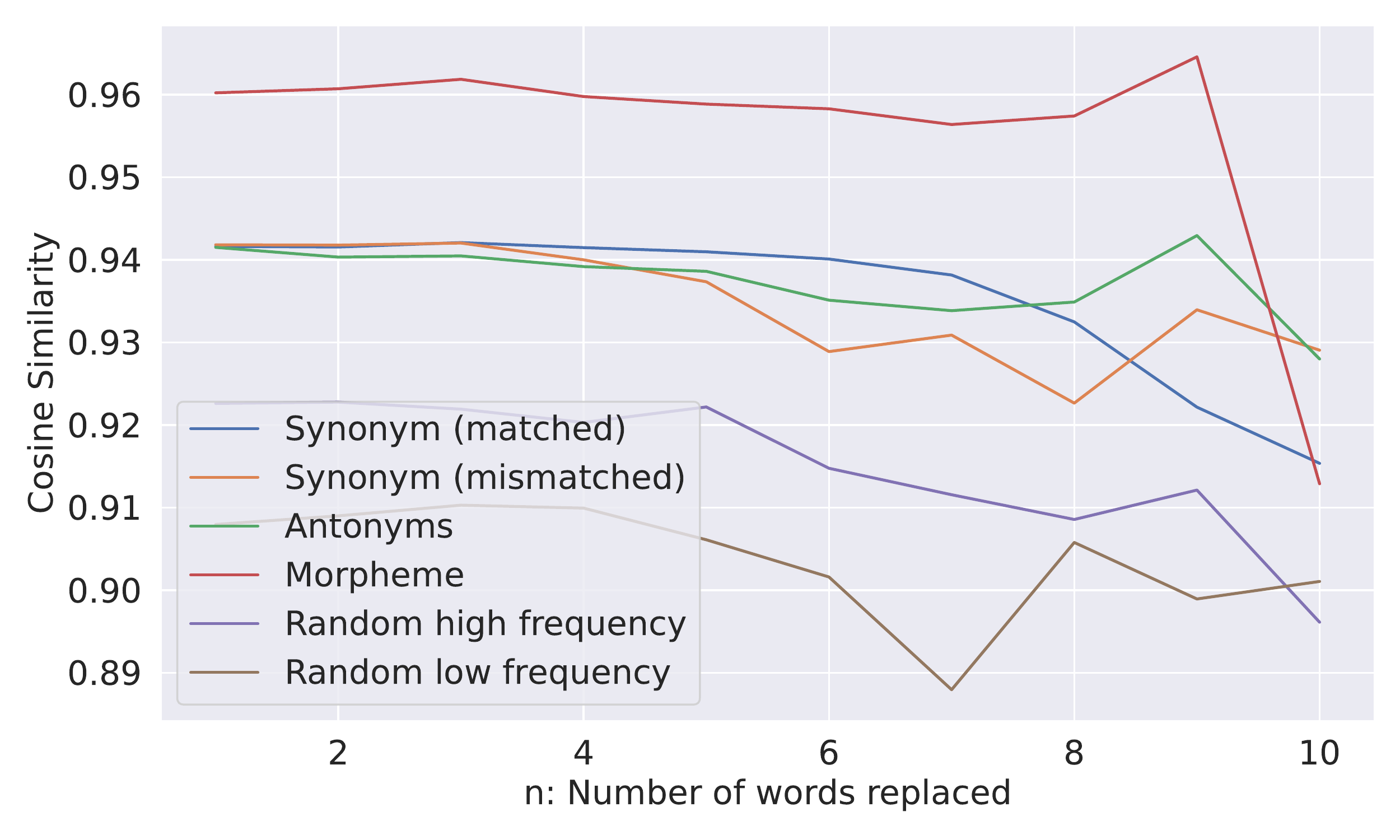}
\caption{ 
The USE sentence embedding cosine similarity between $\mathbf{x}_{ori}$ and the series of sentences obtained by replacing words in $\mathbf{x}_{ori}$ with one type of word substitution.
Different from Figure~\ref{fig:win_original.pdf}, we compare $\mathbf{x}_{swap}^{n}$ with $\mathbf{x}_{swap}^{n-1}$ for $n\geq2$.
The sentence embedding is calculated using a 15-word window around the $n$-th substituted word, as in Figure~\ref{fig:win_original.pdf}. 
}
\label{fig:win_no_original.pdf}
\end{figure}

\begin{figure}[t]
\centering
\includegraphics[width=1.0\linewidth]{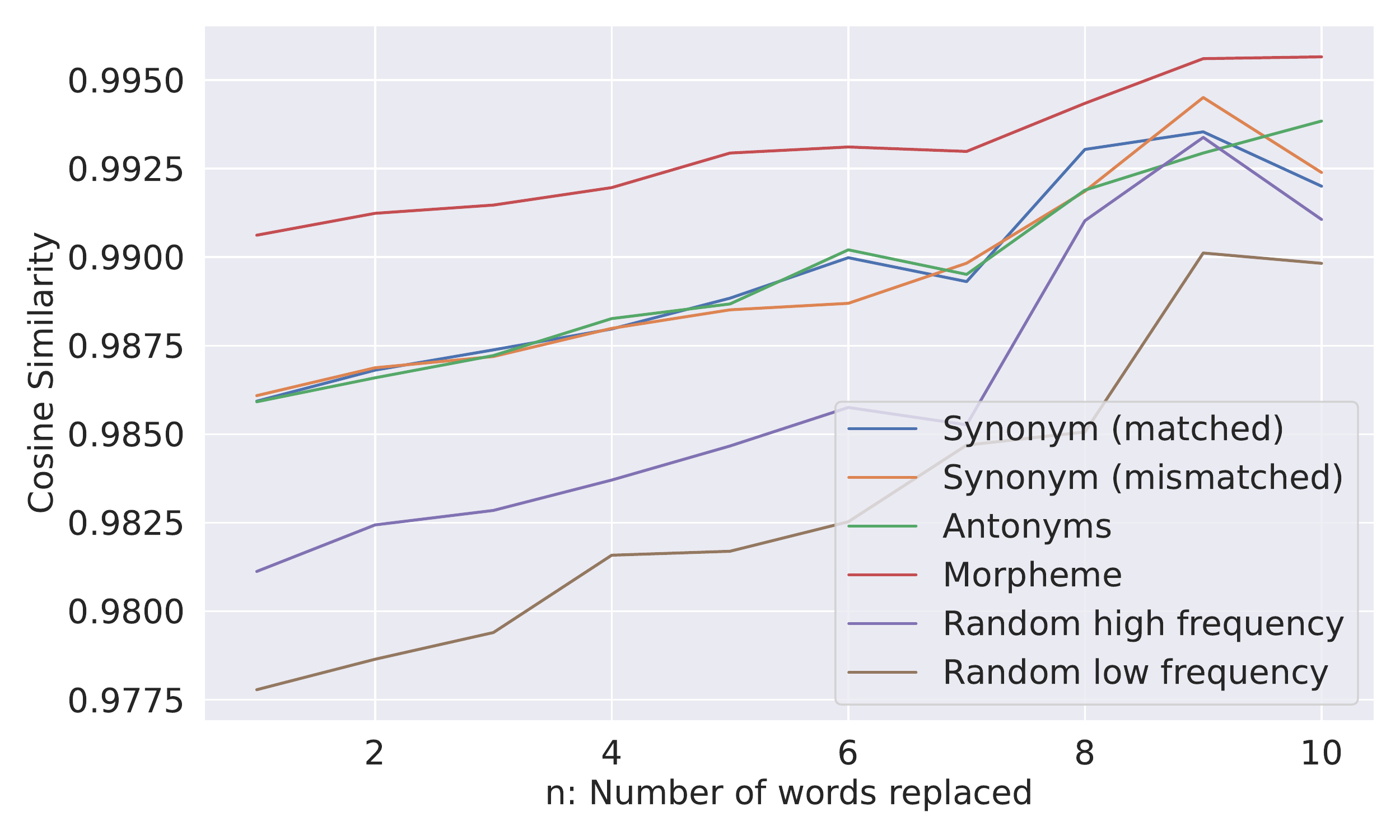}
\caption{ 
The USE sentence embedding cosine similarity between $\mathbf{x}_{ori}$ and the series of sentences obtained by replacing words in $\mathbf{x}_{ori}$ with one type of word substitution.
The sentence embedding similarity shown in this figure is calculated by the whole sentence without windowing and the cosine similarity is calculated between $\mathbf{x}_{swap}^{n}$ and $\mathbf{x}_{swap}^{n-1}$.
}
\label{fig:no_win_no_original.pdf}
\end{figure}

In Figure~\ref{fig:win_no_original.pdf}, we show the result when we we compare $\mathbf{x}_{swap}^{n}$ with $\mathbf{x}_{swap}^{n-1}$ around a 15-word window around the $n-th$ substituted word.
This is adopted in~\citet{garg-ramakrishnan-2020-bae}, according to TextAttack~\citep{morris2020textattack}.
Last, we show the result when we compare $\mathbf{x}_{swap}^{n}$ with $\mathbf{x}_{swap}^{n-1}$ with the whole sentence; this is not used in any previous works, and we include this for completeness of the results.
All the sentence encoders used in Figure~\ref{fig:win_original.pdf},~\ref{fig:no_win_original.pdf},~\ref{fig:win_no_original.pdf},~\ref{fig:no_win_no_original.pdf} are USE.

\subsection{Different Sentence Encoders}
\begin{figure}[t]
\centering
\includegraphics[width=1.0\linewidth]{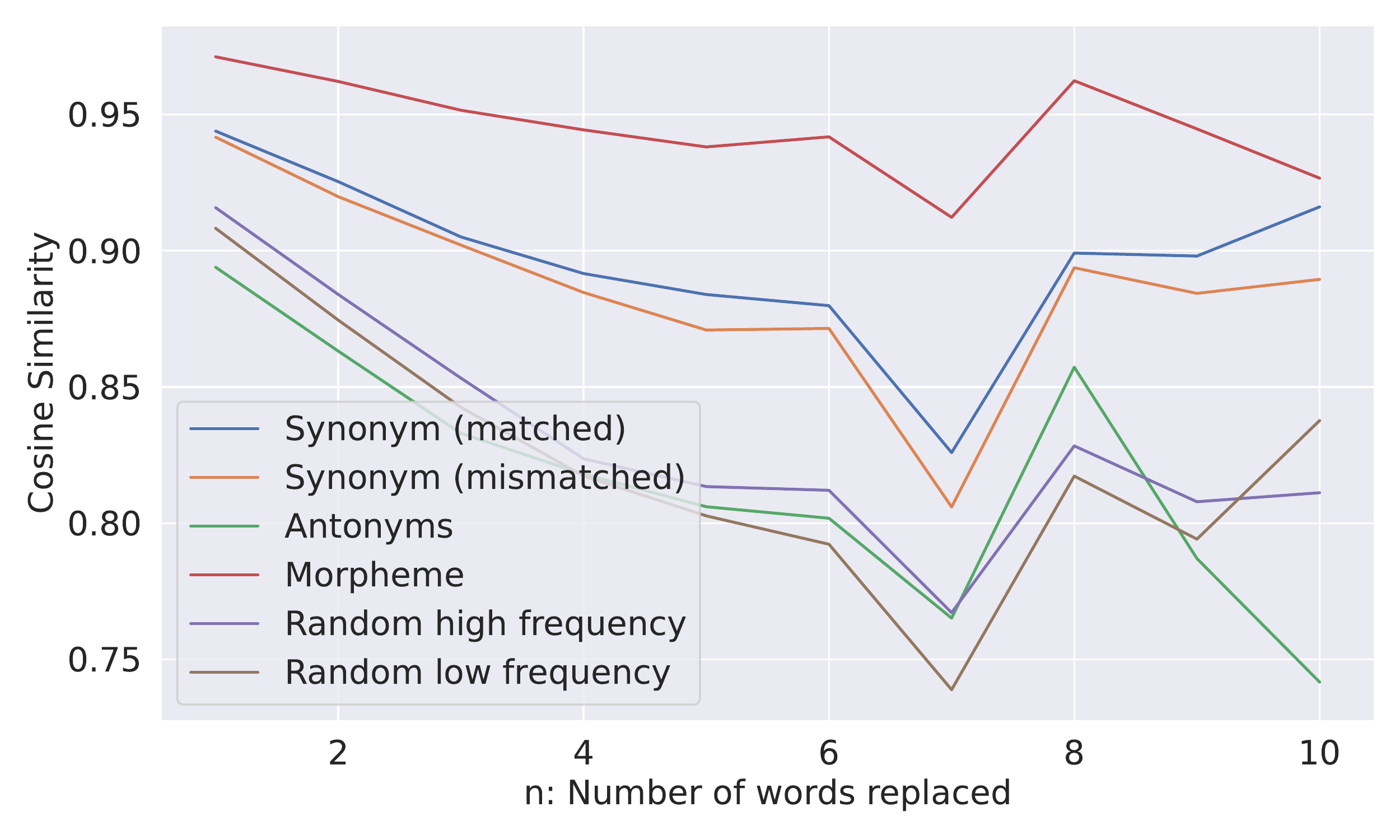}
\caption{ 
Using the DistilBERT fine-tuned on STS-B as the sentence encoder.
Sentence embedding cosine similarity between $\mathbf{x}_{ori}$ and the series of sentences obtained by replacing words in $\mathbf{x}_{ori}$ with one type of word substitution.
}
\label{fig:distilbert.pdf}
\end{figure}

\begin{figure}[t]
\centering
\includegraphics[width=1.0\linewidth]{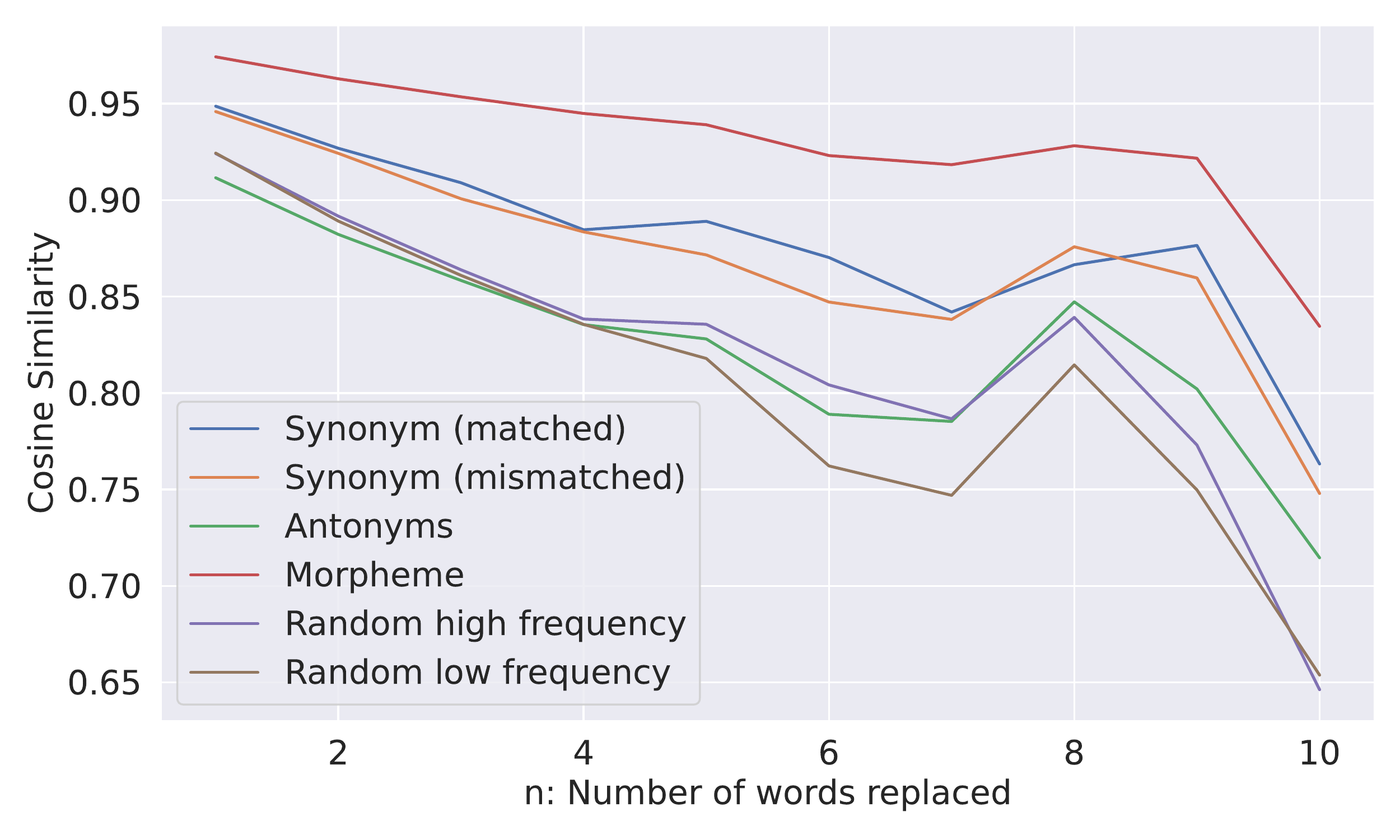}
\caption{ 
Using the DistilRoBERTa fine-tuned on STS-B as the sentence encoder.
Sentence embedding cosine similarity between $\mathbf{x}_{ori}$ and the series of sentences obtained by replacing words in $\mathbf{x}_{ori}$ with one type of word substitution.
}
\label{fig:distilroberta.pdf}
\end{figure}

We show in Figure~\ref{fig:distilbert.pdf} the result when we compare $\mathbf{x}_{swap}^{n}$ with $\mathbf{x}_{ori}$ around a 15-word window around the $n$-th substituted word using a DistilBERT fine-tuned on STS-B, which is the sentence encoder used in~\citet{yoo2021towards}.
Figure~\ref{fig:distilbert.pdf} shows that DistilBERT fine-tuned model better distinguishes between antonyms and synonym swaps, compared with the USE in Figure~\ref{fig:win_original.pdf}.
However, it still cannot distinguish between the matched and mismatched synonym substitutions very well.
Interestingly, this model is flagged as deprecated on \href{https://huggingface.co/sentence-transformers/stsb-distilbert-base}{huggingface} for it produces sentence embeddings of low quality.
We also show the result when we use a DistilRoBERTa fine-tuned on STS-B in Figure~\ref{fig:distilroberta.pdf}.
Interestingly, this sentence encoder can also better distinguish antonym substitutions and synonym substitutions on average.
This might indicate that the models only fine-tuned on STS-B can have the ability to distinguish valid and invalid swaps.

\begin{figure}[t]
\centering
\includegraphics[width=1.0\linewidth]{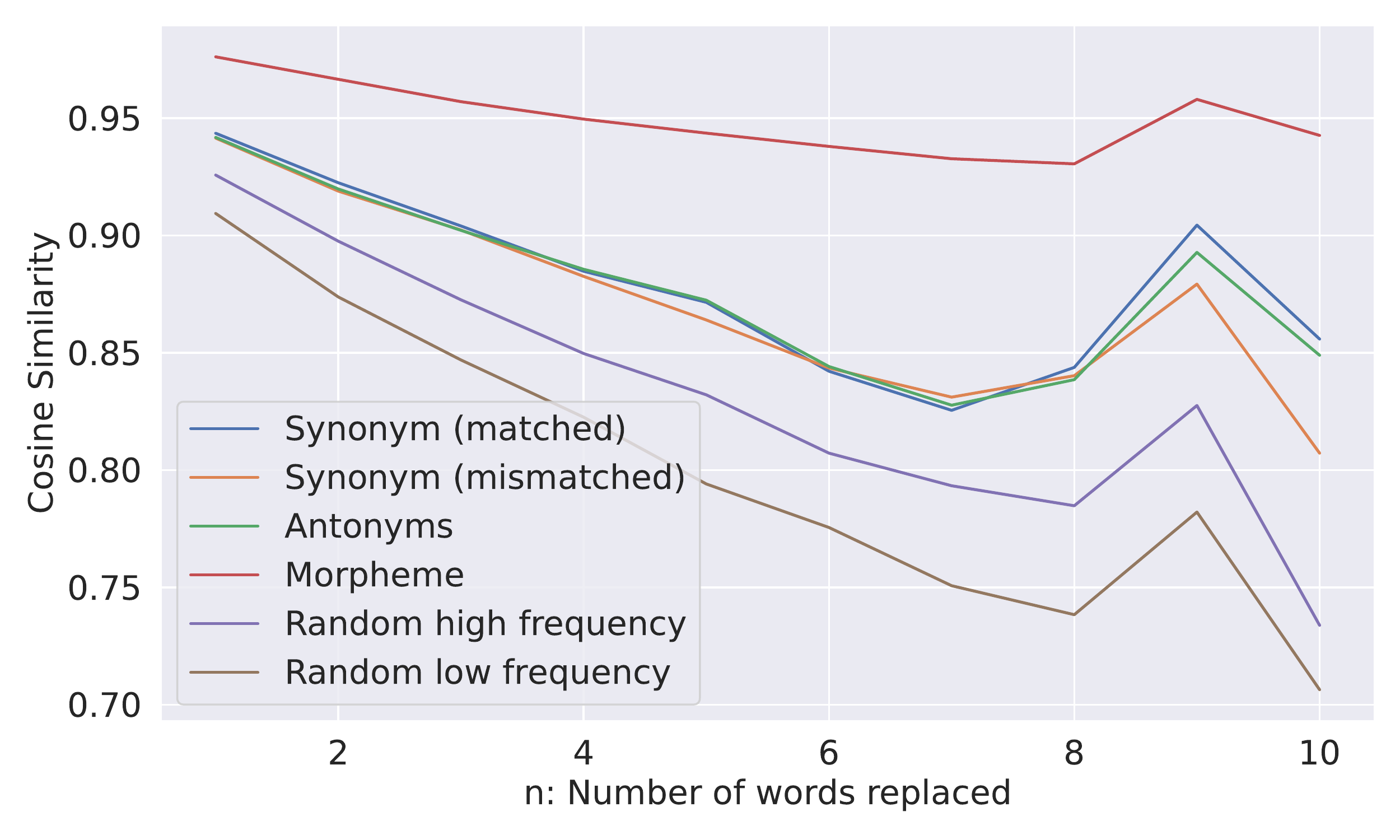}
\caption{ 
The sentence-transformers/all-MiniLM-L12-v2 as the sentence encoder.
Sentence embedding cosine similarity between $\mathbf{x}_{ori}$ and the series of sentences obtained by replacing words in $\mathbf{x}_{ori}$ with one type of word substitution.
}
\label{fig:all-MiniLM-L12-v2.pdf}
\end{figure}

In Figure~\ref{fig:all-MiniLM-L12-v2.pdf}, we show the result when we compare $\mathbf{x}_{swap}^{n}$ with $\mathbf{x}_{ori}$ around a 15-word window around the $n-th$ substituted word using \href{https://huggingface.co/sentence-transformers/all-MiniLM-L12-v2}{sentence-transformers/all-MiniLM-L12-v2}.
This model has 110M parameters and is the 4th best sentence encoder in the pre-trained models on \href{https://www.sbert.net/docs/pretrained_models.html}{sentence-transformer} package~\citep{reimers-2019-sentence-bert}.
It is trained on 1 billion text pairs.
We report the result when using this sentence encoder because it is the best model that is smaller than USE, which has 260M parameters.
We can see that the trend in Figure~\ref{fig:all-MiniLM-L12-v2.pdf} highly resembles that in Figure~\ref{fig:win_original.pdf}, indicating that even a very strong sentence encoder is not suitable to be used as a constraint in SSAs.

\begin{figure}[t]
\centering
\includegraphics[width=1.0\linewidth]{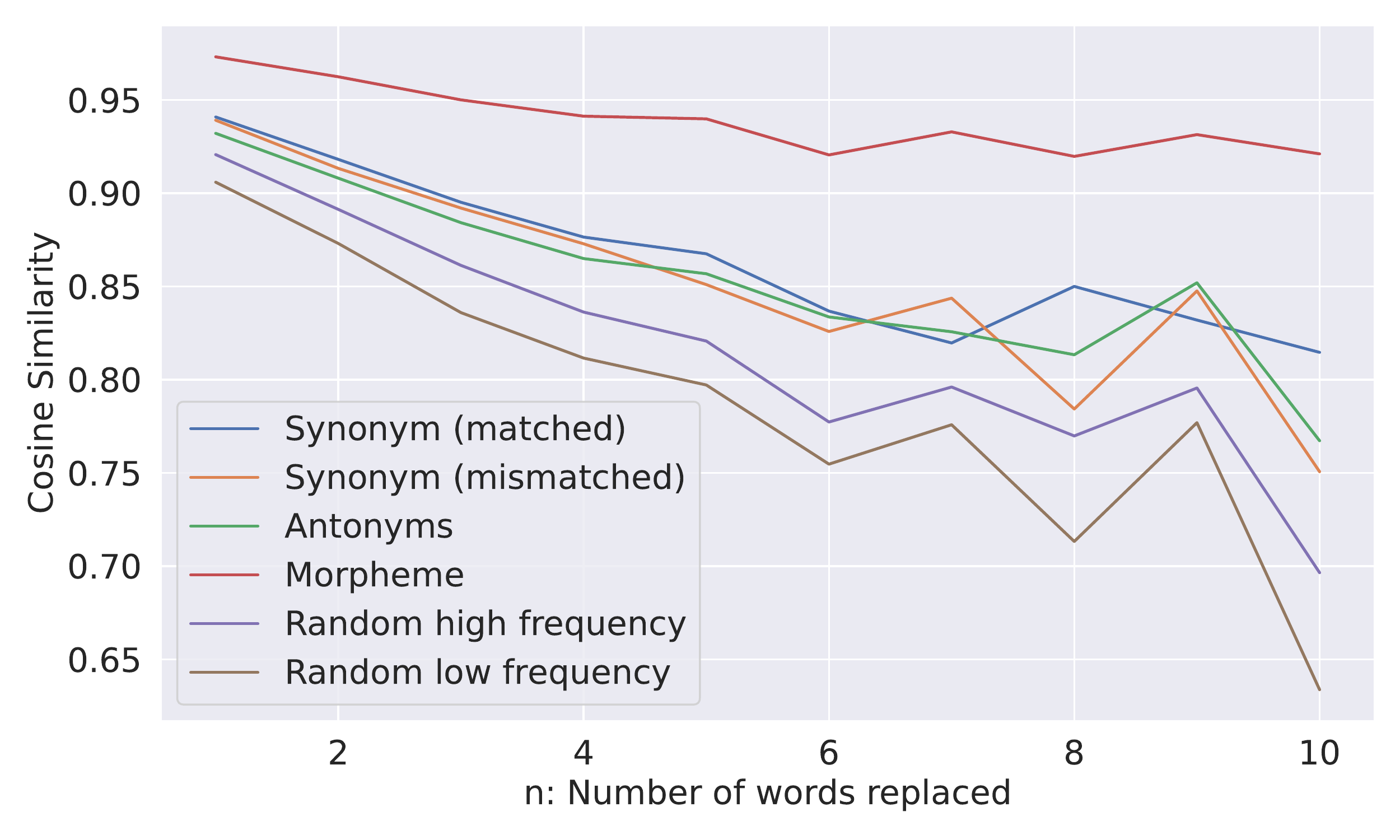}
\caption{ 
The sentence-transformers/all-mpnet-base-v2 as the sentence encoder.
Sentence embedding cosine similarity between $\mathbf{x}_{ori}$ and the series of sentences obtained by replacing words in $\mathbf{x}_{ori}$ with one type of word substitution.
}
\label{fig:all-mpnet-base-v2.pdf}
\end{figure}

We also include the result when we use the best sentence encoder on \href{https://www.sbert.net/docs/pretrained_models.html}{sentence-transformer} package, the \href{https://huggingface.co/sentence-transformers/all-mpnet-base-v2}{all-mpnet-base-v2}.
It has 420M parameters.
The result is in Figure~\ref{fig:all-mpnet-base-v2.pdf}, and it is obvious that it is still quite impossible to use this sentence encoder to filter invalid swaps.

\section{Statistics of Other Victim Models and Other Datasets}
\label{appendix:Statistics of Other Victim Models and Other Datasets}
In this section, we show some statistics on adversarial samples in the datasets generated by~\citet{yoo-etal-2022-detection}.
The main takeaway in this is part is: Our observation in Section~\ref{subsection:Problems with Current Transformation Methods} holds across different types of victim models (LSTM, CNN, BERT, RoBERTa), different SSAs, and different datasets.
\subsection{Proportion of Different Types of Word Replacement}
\label{appendix:Proportion of Different Types of Word Replacement}
First, we show how different word substitution types consist of the adversarial samples of AG-News.
We show the result of four models and four SSAs in Table~\ref{tab:AG-News, PWWS, attack statistics},~\ref{tab:AG-News, TextFooler, attack statistics},~\ref{tab:AG-News, BAE, attack statistics},~\ref{tab:AG-News, TextFooler-Adj, attack statistics}.
This is done by a similar procedure as in Section~\ref{subsection:PWWS Experiments}.

\begin{table*}[t]
    \centering
    \begin{tabular}{|c|c|c|c|c|c|}
    \hline
        Model & Matched sense & Mismatched sense & Morphological & Antonym & Others\\
        \hline
        CNN & 5449 (16.8\%) & 23727 (73.2\%) & 788 (2.43\%) & 0 (0.0\%) & 2434 (7.51\%)\\
        LSTM & 5185 (15.7\%) & 24621 (74.5\%) & 788 (2.38\%) & 0 (0.0\%) & 2467 (7.46\%)\\
        BERT & 4319 (16.2\%) & 19467 (73.2\%) & 1026 (3.86\%) & 0 (0.0\%) & 1788 (6.72\%)\\
        RoBERTa & 5057 (16.3\%) & 21741 (70.2\%) & 1253 (4.05\%) & 0 (0.0\%) & 2905 (9.38\%)\\
        \hline
    \end{tabular}
    \caption{Attack statistics of other models on AG-News.
    The SSA use to attack the models is PWWS.}
    \label{tab:AG-News, PWWS, attack statistics}
\end{table*}

\begin{table*}[t]
    \centering
    \begin{tabular}{|c|c|c|c|c|c|}
    \hline
        Model & Matched sense & Mismatched sense & Morphological & Antonym & Others\\
        \hline
        CNN & 319 (0.891\%) & 897 (2.5\%) & 1464 (4.09\%) & 0 (0.0\%) & 33138 (92.5\%)\\
        LSTM & 304 (0.752\%) & 1125 (2.78\%) & 1662 (4.11\%) & 0 (0.0\%) & 37350 (92.4\%)\\
        BERT & 399 (0.806\%) & 1632 (3.3\%) & 2471 (4.99\%) & 0 (0.0\%) & 45008 (90.9\%)\\
        RoBERTa & 391 (0.783\%) & 1613 (3.23\%) & 2276 (4.56\%) & 2 (0.004\%) & 45656 (91.4\%)\\
        \hline
    \end{tabular}
    \caption{Attack statistics of other models on AG-News.
    The SSA use to attack the models is TextFooler.}
    \label{tab:AG-News, TextFooler, attack statistics}
\end{table*}

\begin{table*}[t]
    \centering
    \begin{tabular}{|c|c|c|c|c|c|}
    \hline
        Model & Matched sense & Mismatched sense & Morphological & Antonym & Others\\
        \hline
        CNN & 34 (1.21\%) & 73 (2.6\%) & 232 (8.25\%) & 5 (0.178\%) & 2468 (87.8\%)\\
        LSTM & 30 (0.998\%) & 62 (2.06\%) & 234 (7.78\%) & 7 (0.233\%) & 2674 (88.9\%)\\
        BERT & 21 (0.88\%) & 39 (1.6\%) & 184 (7.7\%) & 8 (0.34\%) & 2128 (89.4\%) \\
        RoBERTa &25 (0.755\%) & 61 (1.84\%) & 304 (9.18\%) & 6 (0.181\%) & 2914 (88.0\%)\\
        \hline
    \end{tabular}
    \caption{Attack statistics of other models on AG-News.
    The SSA use to attack the models is BAE.}
    \label{tab:AG-News, BAE, attack statistics}
\end{table*}

\begin{table*}[t]
    \centering
    \begin{tabular}{|c|c|c|c|c|c|}
    \hline
        Model & Matched sense & Mismatched sense & Morphological & Antonym & Others\\
        \hline
        CNN & 65 (3.86\%) & 176 (10.5\%) & 706 (42.0\%) & 0 (0.0\%) & 735 (43.7\%)\\
        LSTM & 70 (3.9\%) & 208 (11.6\%) & 698 (38.9\%) & 0 (0.0\%) & 820 (45.7\%)\\
        BERT & 53 (4.32\%) & 118 (9.62\%) & 530 (43.2\%) & 0 (0.0\%) & 526 (42.9\%)\\
        RoBERTa & 59 (4.21\%) & 137 (9.79\%) & 581 (41.5\%) & 0 (0.0\%) & 623 (44.5\%)\\
        \hline
    \end{tabular}
    \caption{Attack statistics of other models on AG-News.
    The SSA use to attack the models is TextFooler-Adj.}
    \label{tab:AG-News, TextFooler-Adj, attack statistics}
\end{table*}

\subsection{Statistics of Different Datasets}
\label{appendix:Statistics of Different Datasets}
In this section, we show the statistics of types of word substitution of another two datasets in~\citet{yoo-etal-2022-detection}.
The result is in Table~\ref{tab:TextFooler BERT, attack statistics}.
Clearly, our observation that valid word substitutions are scarce can also be observed in both SST-2 and IMDB.

\begin{table*}[t]
    \centering
    \begin{tabular}{|c|c|c|c|c|c|}
    \hline
        Model & Matched sense & Mismatched sense & Morphological & Antonym & Others\\
        \hline
        SST-2 & 34 (0.945\%) & 118 (3.28\%) & 206 (5.72\%) & 0 (0.0\%) & 3241 (90.1\%)\\
        IMDB & 1881 (1.43\%) & 4825 (3.66\%) & 8708 (6.6\%) & 21 (0.0159\%) & 116479 (88.3\%)\\
        \hline
    \end{tabular}
    \caption{Attack statistics of other BERT fine-tuned on other datasets.
    The SSA use to attack the models is TextFooler.}
    \label{tab:TextFooler BERT, attack statistics}
\end{table*}

\begin{figure}[t]

\centering
\includegraphics[width=1.0\linewidth]{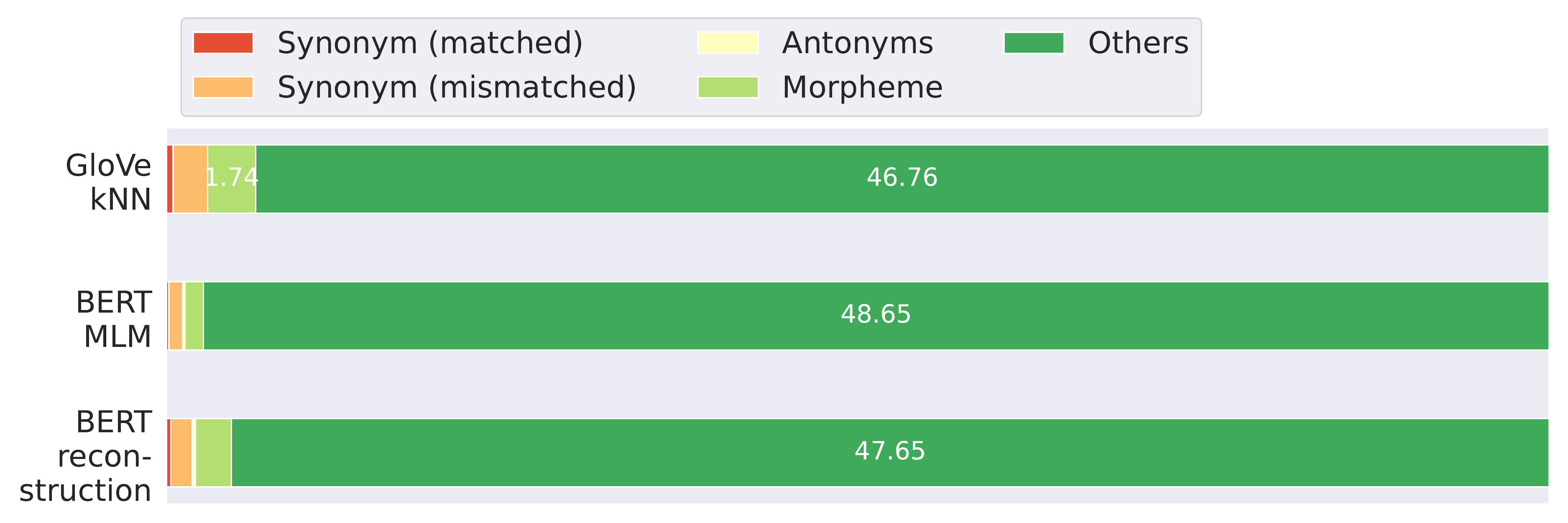}

\caption{ 
The average words of different substitution types in the candidate word set with 50 words for each transformation.
If the average number of words of a substitution type is less than 1.7, we do not show the average number in the bar.
}
\label{fig:bar_50.pdf}
\end{figure}

\end{document}